\PassOptionsToPackage{numbers,compress}{natbib}
\documentclass{article}

\usepackage[preprint,dblblindworkshop]{neurips_2026}

\workshoptitle{AXIOM: Foundations of Efficient Deep Learning}

\usepackage[utf8]{inputenc} %
\usepackage[T1]{fontenc}    %
\usepackage{hyperref}       %
\usepackage{url}            %
\usepackage{booktabs}       %
\usepackage{amsfonts}       %
\usepackage{nicefrac}       %
\usepackage{microtype}      %
\usepackage{xcolor}         %

\usepackage{graphicx}
\usepackage{subfigure}
\usepackage{amsmath}
\usepackage{enumitem}

\title{Functional Degeneracy in Neural Networks: \\Measurement and Pruning}

\author{%
 Maria Matveev$^{1, 2}$
    \quad
Pascal Esser$^{1, 2}$\\
   \quad
  \textbf{Ayush Bharadwaj}$^{3}$
   \quad
  \textbf{Lucius Bushnaq}$^{4}$
  \quad
  \textbf{Gitta Kutyniok}$^{1, 2, 5, 6}$  \vspace{.1cm}\\
    $^1$Department of Mathematics, LMU Munich  $\quad$
    $^2$Munich Center for Machine Learning (MCML)\\
    $^3$Independent $\quad$
    $^4$Goodfire AI $\quad$
    $^5$Department of Physics and Technology, University of Tromsø  \\
     $^6$Institute for Robotics and Mechatronics, DLR-German Aerospace Center  $\qquad$
}

\begin{document}
\maketitle

\begin{abstract}
A central question in modern machine learning is how much a trained model can be compressed without changing its behavior, to reduce the memory, compute and energy required to deploy it. 
To study this, we quantify functional degeneracy through the behavioral recovery rank, the number of leading behavioral-Hessian eigendirections required to recover a model's performance to a given tolerance. Using it as a geometric benchmark for compression, we find that the rank flattens once the task is saturated, while the number of parameters retained by one-shot pruning keeps growing. This gap suggests that functional degeneracy is distributed across parameter directions and not exposed by individual weights or neurons.
\end{abstract}

\section{Introduction}

Trained neural networks exhibit substantial functional redundancy: many parameter directions can be perturbed or removed with little effect on the function realized by the network \cite{grigsby23hidden,lau2025llc,lecun1989obd,Han2015Learning}. To quantify the extent of this redundancy, we ask:\\
\centerline{\emph{Q1: How many parameter directions are required to preserve the learned function?}}

A high degree of degeneracy indicates that the learned function is insensitive to many parameter directions, placing the solution in a broad, approximately flat region associated with generalization and stable optimization \cite{Goodfellow-et-al-2016,Hochreiter97,keskar2017on,garipov2018losssurfacesmodeconnectivity,NeyshaburBMS17}. This redundancy also provides the basis for model compression: if many parameters have little influence on the learned function, retaining only a suitable subset may preserve its approximate performance while reducing the memory, compute, and energy required for deployment \cite{Han2015DeepCC,Frankle2018TheLT}. Although model sizes continue to grow, achievable compression is still often measured post-hoc by applying a particular method \cite{blalock2020pruning}. It therefore becomes increasingly important to assess the \emph{functional degeneracy of a model}, and to determine how effectively existing compression methods exploit it. This motivates our second question:\\
\centerline{\emph{Q2: How much of this redundancy do we actually exploit through pruning?}}

\textbf{Contributions.} 
To shed light on these questions, we make the following contributions:

    \emph{(i)} We introduce the \emph{behavioral recovery rank}, the number of leading behavioral Hessian eigendirections needed to recover a model's performance to a given tolerance. This provides a geometrically motivated way to measure approximate functional degeneracy of a model (Sec.~\ref{sec:methodology}, App.~\ref{app:recovery_variants}-\ref{app:behavioral_hessian}). 
    
\begin{figure}[t]
    \centering
    \subfigure[Test MSE for projecting on varying number of eigendirections.]{
    \includegraphics[height=3.1cm,trim=2mm 1mm 3mm 1mm, clip]{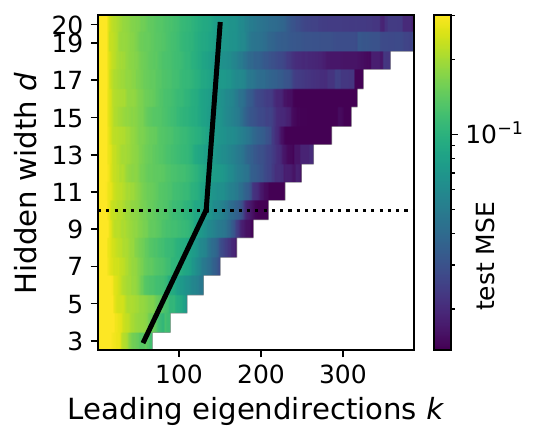}
    }
    \hfill
    \subfigure[Behavioral recovery rank extracted from the projection curves, mean and std. over $5$ runs.\protect\footnotemark]{
       \includegraphics[height=3.1cm,trim=2.5mm 1mm 3.5mm 1mm, clip]{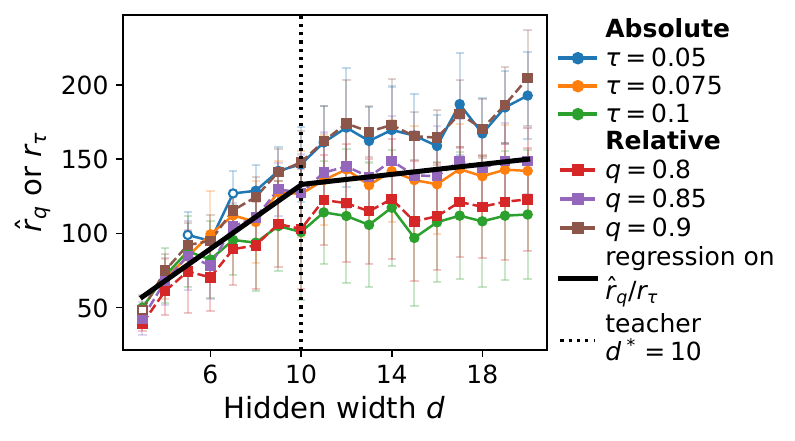} 
      }
      \hfill
          \subfigure[Pruning and behavioral recovery rank at a fixed loss threshold.]{
    \includegraphics[height=3.1cm,trim=2mm 1mm 2mm 1mm, clip]{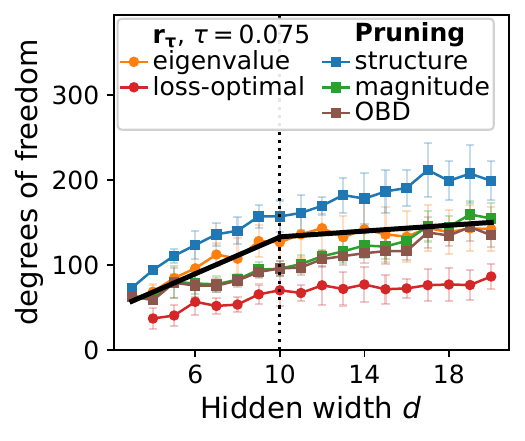}
    }
    \caption{Task-saturated and capacity-limited regimes for non-linear teacher-student regression. Trend-lines (piecewise regression) of both regimes are given in black.}
    \label{fig:2regimes}
\end{figure}

    \emph{(ii)} We use controlled nonlinear teacher-student experiments, where task complexity is known, to show that the behavioral recovery rank separates \emph{capacity-limited} from \emph{task-saturated regimes}. The behavioral recovery rank grows with student width and, at moderate performance tolerances, flattens once the student exceeds the teacher width (Fig.~\ref{fig:2regimes}(a)-(b)). On MNIST, it keeps growing with width, roughly in proportion to the parameter count, and remains a small fraction of it (Fig.~\ref{fig:mnist}).
    
    \emph{(iii)} We find that several one-shot pruning methods do not exhibit such flattening, i.e. the parameters they retain keep growing past the teacher's width (Fig.~\ref{fig:2regimes}(c)), and 
    exceed the behavioral recovery rank under loss-optimal ordering of directions.

Together, these findings suggest that functional degeneracy beyond task saturation is not exposed by weight- or neuron-level structure, and is therefore only partially accessible to standard pruning.

\addtocounter{footnote}{-1}

\footnotetext{Hollow markers indicate that the mean is over fewer than five seeds, as the tolerance is not reached for at least one seed.}

\textbf{Related Work.} Our questions are motivated by several works on degeneracies in neural network parameterizations. \citet{grigsby23hidden} characterize \emph{exact} degeneracies, and follow-up work formalizes this perspective through functional dimension, given by the rank of the Jacobian of the realization map on a finite-probe\footnote{A \emph{probe} set is a finite i.i.d.\ reference sample from the data distribution (e.g., training or validation data).}
set \cite{grigsby2025functional}.
However, we argue that, for understanding task performance, the definition of \emph{exact symmetry is too narrow}: a parameter direction can have only a small effect on the realized behavior of the model without being \emph{exactly} redundant. This viewpoint is reinforced by the works on \emph{intrinsic dimension}, showing that many tasks can be solved in subspaces far smaller than the parameter space \cite{li2018intrinsic,aghajanyan2021intrinsic}, and by analyses of the \emph{loss Hessian}, whose spectrum typically shows a large near-zero bulk alongside few outliers \cite{sagun2018empirical,papyan2018fullspectrum,singh2024closed,singh2021analytic,noci2024superconsistency}. 
\emph{Pruning} removes weights or neurons from a trained network while preserving its function. \emph{Unstructured} methods rank individual weights by magnitude \cite{Han2015Learning} or by a local quadratic model of the loss \cite{lecun1989obd}. \emph{Structured} methods remove whole neurons, and hence speed up computation, ranking them by norm or by an estimated change in the loss \cite{li2017pruning,molchanov2019importancepruning}. Both are typically combined with retraining and evaluated by the fraction of parameters that can be removed at a given accuracy \cite{blalock2020pruning}. We expand on these works in App.~\ref{sec: app related work}.

\section{Methodology} \label{sec:methodology}

\textbf{Quantifying degeneracy: Behavioral recovery rank.}
We first equip the neighborhood of the parameters at a checkpoint during training with a geometry that tracks changes in functional behavior, and then quantify how many of the  directions it identifies as behaviorally most relevant are needed to retain a model's performance.
Formally, for a model $f_\theta$  parameterized by $\theta\in\mathbb R^p$,  following \citet{bushnaq2024degeneracy}, we consider the \emph{behavioral loss}\footnote{For notational simplicity we only consider the MSE as a loss; however, the setup generalizes to other loss functions.} for two sets of parameters $\tilde\theta$ and $\theta_\ast$
\begin{equation}
L_{\mathrm{beh}}(\tilde\theta;\theta_\ast,\mathcal P) = (2|\mathcal P|)^{-1} \sum\nolimits_{x\in\mathcal P} \|f_{\tilde\theta}(x)-f_{\theta_\ast}(x)\|^2, \label{eq:intro_beh}
\end{equation}
where $\mathcal P=\{x_i\}_{i=1}^n$ is a probe set, and $\|\cdot\|$ denotes the $\ell_2$ norm. This loss equips the neighborhood of $\theta_\ast$ with a geometry tied directly to preservation of learned behavior: perturbations are small when the predictions $f_{\tilde\theta}(x), x\in \mathcal P$, compared to the reference model's predictions $f_{\theta_\ast}(x)$ are nearly unchanged on the probe set.
This is in contrast to the empirical loss. Instead of comparing model predictions to ground-truth labels, the behavioral loss compares predictions to those of a reference checkpoint and can therefore be interpreted as a form of local self-distillation, with the checkpoint itself acting as the teacher \cite{hinton2015distilling}. By construction, $\theta_\ast$ is a global minimizer of $L_{\mathrm{beh}}(\tilde\theta;\theta_\ast,\mathcal P)$, so the first-order term in its Taylor expansion around $\theta_\ast$ vanishes. Consequently, the leading nontrivial local object is the behavioral Hessian,
$
H_{\mathrm{beh}}(\theta_\ast)
:=
\nabla^2_{\tilde\theta}L_{\mathrm{beh}}(\tilde\theta;\theta_\ast,\mathcal P)\big|_{\tilde\theta=\theta_\ast},
$
whose structure we discuss further in App.~\ref{app:behavioral_hessian}. A second-order Taylor expansion yields 
$
L_{\mathrm{beh}}(\theta_\ast+\delta;\theta_\ast,\mathcal P)=
\frac12 \delta^\top H_{\mathrm{beh}}(\theta_\ast)\delta 
+ o(\|\delta\|^2)$ 
for 
$\delta\in\mathbb R^p.
$
Thus, the leading eigenvectors of $H_{\mathrm{beh}}(\theta_\ast)$ identify perturbation directions that induce the largest changes in model behavior on the probe set. Formally, let $v_1, v_2, \dots$ denote the eigenvectors of $H_{\mathrm{beh}}(\theta_\ast)$ with eigenvalues $\lambda_1\ge\lambda_2\ge\dots$, and let $\Pi_k$ denote the projection onto the span of $v_1,\dots,v_k$, which is unique when $\lambda_k>\lambda_{k+1}$. 
Given a reference point $\theta_{\mathrm{ref}}$, we reconstruct the learned displacement  with respect to $\theta_\text{ref}$ by 
$\hat\theta_k
=
\theta_{\mathrm{ref}}
+
\Pi_k(\theta_\ast-\theta_{\mathrm{ref}}),
$
where we use the reference point $\theta_{\mathrm{ref}}=0$ in the teacher-student experiments and the initialization $\theta_{\mathrm{ref}}=\theta_0$ for MNIST. 
We define the \emph{behavioral recovery rank} as the smallest number of leading eigendirections $k$ such that the projected checkpoint $f_{\hat\theta_k}$ recovers a given performance level.\!\footnote{We only define $\hat r_q$ if $\mathcal M(\theta_{\mathrm{ref}})\neq\mathcal M(\theta_\ast)$. For the accuracy on MNIST, where higher is better, we use $r_\tau=\min\{k:\mathcal M(\hat\theta_k)\ge\tau\}$.
The definition of $\hat r_q$ applies unchanged.
App.~\ref{app:recovery_variants} provides a reordered variant.} 
For a ``lower-is-better'' metric $\mathcal M$, such as the empirical training/test loss, with tolerance $\tau>0$ and  $q\in(0,1]$, we define 
\begin{align*}
    \text{Relative: }  \hat r_q = \min_{k\in[p]}\Big\{ k: \frac{\mathcal M(\theta_{\mathrm{ref}})-\mathcal M(\hat\theta_k)}{ \mathcal M(\theta_{\mathrm{ref}})-\mathcal M(\theta_\ast) } \ge q \Big\},\quad
    \text{Absolute: }  r_\tau = \min_{k \in [p]}\big\{ k: \mathcal M(\hat\theta_k)\le \tau \big\}. 
\end{align*}
Intuitively, a small behavioral recovery rank means that the trained model's behavior can be recovered from a low-dimensional part of the trained solution; if this rank grows much more slowly than parameter count, additional parameters mainly enlarge the degeneracy.\\ 
In our main experiments, we use the train set as the probe set (subset of $10{,}000$ images for MNIST) to compute the behavioral loss in Eq.~\eqref{eq:intro_beh} at a trained model checkpoint $\theta_\ast$. We ablate this choice in App.~\ref{app:further_experimental_results}. We estimate the top-$k$ eigenspaces of the Hessian using matrix-free Lanczos iteration; see App.~\ref{app:implementation} for details and a link to the code.

\textbf{Pruning baselines.}
We apply three one-shot baselines to the same checkpoints. \emph{Magnitude pruning} removes the globally smallest parameters \cite{Han2015Learning}; \emph{structural pruning} removes neurons greedily with respect to $\mathcal{M}$  on the training data. \emph{Optimal Brain Damage} (OBD) \cite{lecun1989obd} ranks parameters by a saliency computed from the diagonal of the behavioral Hessian. 
For each we report the smallest number of retained parameters for which the model performance $\mathcal{M}$ stays within tolerance $\tau$, counting for structural pruning all parameters attached to surviving neurons. This way, both quantities count degrees of freedom retained to meet the tolerance $\tau$, in the coordinate basis for pruning and in the behavioral-Hessian eigenbasis for $r_\tau$.  For comparability with the projection experiments, we do not retrain. In App.~\ref{app:pruning_refit} we report a \emph{refitted} variant and apply the same pipeline to the teacher itself.

\textbf{Model and data.} We study degeneracy in two settings of varying difficulty. Firstly, in nonlinear teacher-student regression, we fix a teacher model and vary student width from under- to over-complete, which allows for a systematic control of the network capacity relative to the data complexity. Secondly, we complement these experiments with width-scaled MLPs on a standard classification task (MNIST \cite{6296535}). For all experiments, we train with Adam without weight decay \cite{Kingma2014AdamAM} for a fixed number of steps. We provide details on both setups in App.~\ref{app:dataset}. 

\section{Results and Discussion}\label{sec: results}

\begin{figure}[t]
    \centering
    \subfigure[Projection accuracy, 
    compared to final accuracy (horizontal lines).]{
    \includegraphics[height=3.1cm,trim=2.5mm 1mm 2.5mm 1mm, clip]{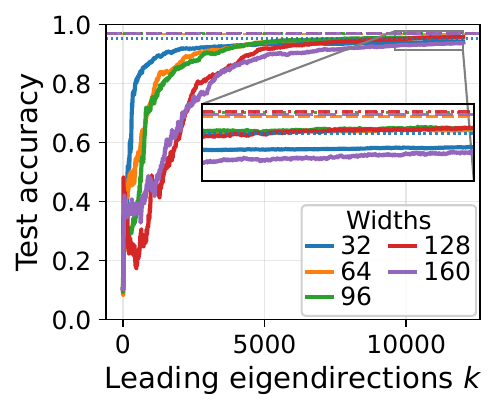}
    }
    \hfill
    \subfigure[Leading eigendirections $k$ required to reach an accuracy threshold $\tau$ or $q$, for different widths.]{
       \includegraphics[height=3.1cm,trim=2.5mm 1mm 4mm 1mm, clip]{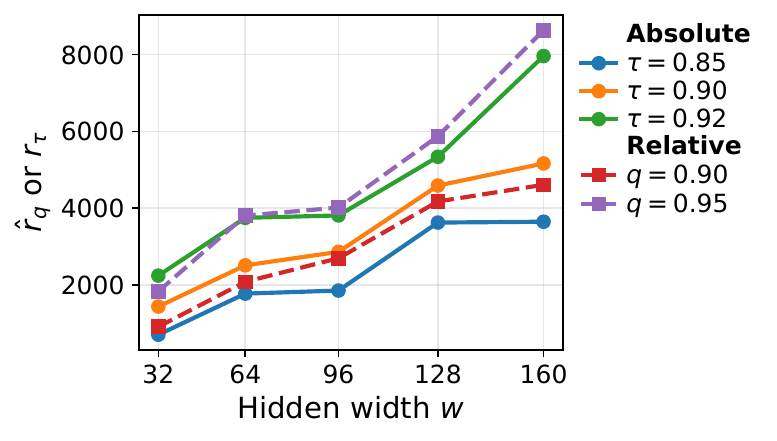}
    }
    \hfill \subfigure[Comparison of behavioral recovery rank to three pruning baselines.]{
       \includegraphics[height=3.1cm,trim=2.5mm 1mm 2.5mm 1mm, clip]{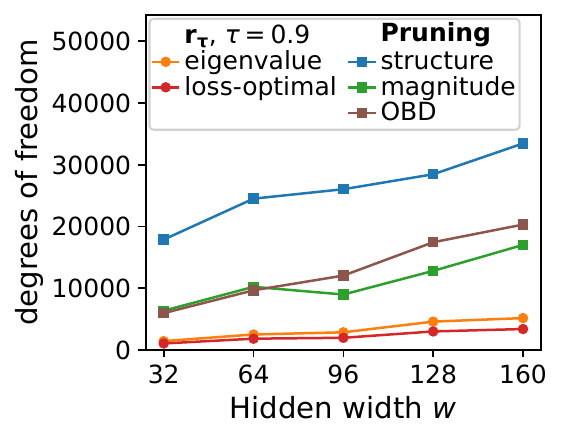}
    }
    \caption{On MNIST, wider models need more directions to recover performance, indicating additional behaviorally relevant directions roughly in proportion to the parameter count and width.}
    \label{fig:mnist}
\end{figure}

\textbf{Q1: The task-saturated and capacity-limited regimes.}
The teacher-student setting allows us to observe the transition between two regimes directly, because task complexity is fixed by the teacher width, giving us a priori control over model capacity relative to the task. 
While the student is \emph{capacity-limited} (narrower than the teacher), added parameters  improve 
performance gradually as a large fraction of the available directions contributes to the learned behavior as shown in Fig.~\ref{fig:2regimes}(b). 
As the student matches or exceeds the teacher width, the task becomes \emph{saturated}: the recovery curves in Fig.~\ref{fig:2regimes}(a) become increasingly steep, only a limited number of leading directions is needed to approach the final performance, and additional directions yield diminishing improvements. The piecewise regression in Fig.~\ref{fig:2regimes}(b) reveals that the growth of the behavioral recovery rank for these threshold levels slows from $10.8$ to $1.7$ directions per added neuron ($21$ parameters). Absolute and relative recovery criteria exhibit the same qualitative transition, except for tight tolerances (Fig.~\ref{fig:tight}(a)).
For MNIST, the task complexity is unknown, and the operating regime must therefore be inferred from the scaling of the rank. Fig.~\ref{fig:mnist}(a)-(b) shows that wider models require more leading eigendirections to reach the same accuracy, although their final accuracies differ little. The recovery rank grows roughly
in proportion to the parameter count and width under both absolute and relative criteria. This is consistent with a capacity-limited regime, 
in the sense that width introduces additional behaviorally relevant directions. 

\textbf{Q2: Redundancy accessible to pruning.}
We now use the behavioral recovery rank as a geometric benchmark for assessing how much of the identified redundancy is exploited by pruning. Fig.~\ref{fig:2regimes}(c) compares the behavioral recovery rank with the three pruning methods at a fixed performance threshold. Once the student width reaches the teacher width, the behavioral recovery rank slows markedly, whereas the number of parameters retained by the three pruning methods continues to grow. Hence, the additional redundancy arising beyond task saturation is not effectively exposed by these one-shot pruning criteria, which therefore underestimate the degeneracy present in this regime.
\\The red curve in Fig.~\ref{fig:2regimes}(c) reports the same projection experiment but orders directions by $\lambda_j\langle v_j,\theta_\ast-\theta_\text{ref}\rangle^2$ rather than by $\lambda_j$ alone. Under the quadratic model, this ordering minimizes the behavioral loss for a fixed budget of $k$ directions on the probe set. It approximately halves the recovery rank, indicating even greater degeneracy than the eigenvalue ordering reveals, and all three pruning baselines retain more degrees of freedom at every width. Since individual behavioral eigendirections can mix parameters across all hidden units, whereas pruning removes units or weights in the original parameterization, the observed gap suggests that the redundant directions are not aligned with these architectural coordinates. On MNIST, all three baselines retain more parameters than the behavioral recovery rank which keeps growing with the width (Fig.~\ref{fig:mnist}(c)).

\textbf{Discussion.}
The behavioral recovery rank provides a task-dependent distinction between nominal model size and the effective number of parameter directions needed to preserve learned behavior. Pruning operates locally, on the individual weights and units that the architecture provides, whereas each behavioral eigendirection combines all parameters. Although low-dimensional, this eigenbasis representation is not directly feasible for compression, since each basis element is as large as the dense model. The behavioral recovery rank indicates instead what a function-adapted basis could remove. Consistent with this, closed-form output refitting recovers much of the gap (App.~\ref{app:pruning_refit}). Whether such a basis is also inexpensive to describe, for instance via structured families such as layerwise or blockwise transforms, which are already used to approximate curvature in second-order optimizers \cite{gupta18shampo,martens2015kfac}, and how to utilize it for compression remain open.  \\ 
Because the behavioral recovery rank flattens  while the parameter count and the parameters retained by pruning continue to grow, compression measured only relative to the original parameter count may understate the redundancy that remains once task complexity is saturated. This opens an avenue towards predictive theory, since degeneracy fixed by the task could be measured on a small model and used to estimate what compression can achieve on larger ones.

\textbf{Limitations and future work.} Our experiments cover two small settings where repeated behavioral-Hessian eigensolves remain tractable. Making the behavioral recovery rank operational may require blockwise restriction or probe-set subsampling. Testing whether the regimes persist at scale requires broader architectures and datasets, with ablations over the probe set, reference point and parameterization to distinguish genuine capacity effects from width-dependent optimization and regularization. 
Although the directions of function-preserving rescalings of ReLU units lie in the kernel of $H_{\mathrm{beh}}$, such reparameterizations change its leading eigenspaces (App.~\ref{app:behavioral_hessian_fd}). Moreover, $H_{\mathrm{beh}}$ describes the behavioral loss only locally around $\theta_\ast$, whereas the projections $\hat\theta_k$ for small $k$ can lie far from it.
Our three one-shot pruning baselines may understate compression achievable through retraining or quantization. Finally, a theoretical characterization of the behavioral recovery rank could clarify when these width-scaling regimes persist and support compression or generalization guarantees.

\textbf{Acknowledgments}
  MM, PE and GK acknowledge the support by the Munich Center for Machine Learning (MCML).  MM and GK are supported by the DAAD programme Konrad Zuse Schools of Excellence in Artificial Intelligence, sponsored by the German Federal Ministry of Research, Technology and Space. We thank Sid Baines, George Ingebretsen, and Hernan Iriarte for helpful discussions in the early stages of this work.

\bibliographystyle{plainnat}
\bibliography{sample}

\newpage
\clearpage
\appendix

\section{Ethical Considerations}
\subsection{Impact statement}
This paper presents work whose goal is to advance the field of Machine Learning. There are many potential societal consequences of our work, none of which we feel must be specifically highlighted here.

\subsection{Disclosure of AI use}
We utilized AI tools for support with coding, editing, and ideation. All AI-generated suggestions were critically reviewed and refined by the authors, and we take full responsibility for the work presented.

\section{Extended Related Work Discussion}\label{sec: app related work}

In the following, we discuss the related work directions outlined in the introduction in more detail.

\textbf{Scaling laws} describe how model loss varies with model size, data, and compute \cite{kaplan2020scaling}. \citet{hoffmann2022empirical} show that model size and token count should be scaled together, and \citet{bahri2024explaining} emphasize that different regimes can exhibit different exponents. This motivates studying how the behavioral recovery rank varies with the hidden size of the model. 

\textbf{Singular learning theory} studies model complexity through the local singular structure around a solution \cite{watanabe2009algebraic}. The local learning coefficient (LLC) quantifies how the volume of a low-loss region shrinks as one zooms in around a local minimum \cite{lau2025llc}.  Following this local perspective, we adopt the \emph{behavioral loss} introduced by \citet{bushnaq2024degeneracy}. In their framework, the LLC of the behavioral loss defines an effective parameter count, and the rank of its Hessian gives a tractable lower bound on the corresponding local dimension. They further connect this functional degeneracy to concrete network properties, such as linear dependence in both the activations and the backpropagated gradients, and use the so-called interaction basis to interpret the model independent of parameterization. 

\textbf{Hessian analysis.} Training is often studied through the \textit{Hessian of the training or validation loss}. A recurring phenomenon, observed empirically and analyzed in simplified settings, is that the loss Hessian possesses a spectrum with a large near-zero bulk together with a small set of outliers \cite{sagun2018empirical,papyan2018fullspectrum,singh2024closed}. Analytic work further shows that such rank deficiency can follow from architectural structure itself such as width, depth, and bias configuration \cite{singh2021analytic}. More recently, \citet{noci2024superconsistency} show that under $\mu$P parameterization, certain spectral properties of the loss Hessian remain largely stable across model sizes, a phenomenon they call \emph{super consistency}. Works on the Edge-of-stability further connect the loss-curvature to optimization dynamics via the learning rate \cite{cohen2021gradient}. Together, these results suggest that increasing parameter count need not create proportionally many new relevant \emph{loss curvature} directions. By contrast, our behavioral construction replaces supervised targets by the outputs of a fixed trained checkpoint on a probe set, so that the curvature is induced by these  outputs instead of the labels. This can be viewed as local self-distillation with the checkpoint itself as the teacher \cite{hinton2015distilling}. We discuss the connection of both second-order objects in App.~\ref{app:behavioral_hessian_task}.

\textbf{Intrinsic dimension.} The works on \emph{intrinsic dimension} suggest that once a model family is sufficiently large, the actual dimensionality required to solve a task remains relatively stable \cite{li2018intrinsic}. \citet{aghajanyan2021intrinsic} report similarly low intrinsic dimensionality in language-model fine-tuning. Intrinsic dimension is defined through training in random low-dimensional subspaces, whereas our statistic is computed after training in a Hessian-aligned subspace around a fixed checkpoint. Still, these results provide a clear prior for our setting: the dimension relevant to the learned solution may grow substantially more slowly than the total number of parameters. This is also echoed by the \textit{manifold hypothesis}, which posits that high-dimensional data concentrate near lower-dimensional structure \cite{cayton2005manifoldlearning,whiteley2026statistical}.

\textbf{Functional dimension} measures the local dimension of the realized function class and can be computed
on finite batches via the rank of the evaluation-map Jacobian \cite{grigsby2025functional}. \citet{grigsby23hidden} show that exact degeneracies of ReLU networks can depend on architectural parameters.
The same finite-probe Jacobian also appears in the empirical \emph{Neural Tangent Kernel} theory \cite{jacot2018neural}, its spectrum was analyzed in \cite{fan2020spectra}. However, we use this object post hoc around trained checkpoints, rather than to analyze training dynamics near initialization. We discuss these connections in App.~\ref{app:behavioral_hessian}.

\textbf{Pruning.} Pruning asks which parameters can be removed from a trained network while preserving its function. The classical
saliency methods estimate the effect of removing a weight through a local quadratic model of the loss, either under a diagonal approximation of the Hessian \cite{lecun1989obd} or using its inverse to additionally compensate the remaining weights \cite{hassibi1993obs}. Magnitude pruning replaces this by the far cheaper proxy of weight size \cite{Han2015Learning}, which remains competitive at scale and forms the first stage of standard compression pipelines \cite{Han2015DeepCC}. Unstructured sparsity does not translate into speedups on dense hardware without specialized kernel support. \emph{Structured} pruning methods remove neurons, directly impacting the weight matrix size involved in the computation, by ranking them by norm \cite{li2017pruning} or by a first-order estimate of the induced change in the loss \cite{molchanov2019importancepruning}. These methods are typically combined with retraining \cite{Han2015Learning} and evaluated by what fraction of the parameter count can be removed while still meeting an accuracy threshold \cite{blalock2020pruning}. In either case the criterion is expressed in the coordinate basis of the parameterization, whereas the behavioral recovery rank counts directions in the eigenbasis of the behavioral Hessian, which need not align with any weight or unit. To compare with the projection experiments, we do not retrain and use a greedy pruning approach, as described in Sec.~\ref{sec:methodology}.

\section{Experimental details}\label{app:exp_details}
This section provides the detailed experimental configuration used throughout the paper, including datasets, model architectures, training procedures, and implementation specifics. We describe both the synthetic teacher-student regression setup and the MNIST classification experiments in Sec.~\ref{app:dataset}, together with the optimization settings and computational methods used to analyze the resulting models in Sec.~\ref{app:implementation}.
\subsection{Data, models, training}
\label{app:dataset}
In this section, we introduce the datasets and experimental setups. We describe the data generation process for the synthetic regression experiments in Sec.~\ref{app: data student teacher} as well as the preprocessing and training configuration for MNIST classification in Sec.~\ref{app: data: mnist}.
\subsubsection{Nonlinear teacher-student regression}
\label{app: data student teacher}
In order to systematically control model capacity, we utilize a synthetic teacher-student regression setup with two-layer ReLU MLPs.

For this, we sample the inputs according to 
$
x \sim \mathcal N(0,I_{d_{\rm in}}).
$
The teacher network $f_T$ generates regression targets $y=f_T(x)+\xi$ with $\xi\sim \mathcal N(0,\sigma^2I)$.

Both teacher and student are two-layer ReLU networks, with bias terms.  The biases of the teacher are set to zero, and only its weights are drawn at random. For the experiments in the main paper, the teacher has architecture $10\to 10\to 10$
and the student has architecture $10\to d\to 10$, and the data is noiseless. These choices are ablated in App.~\ref{app:further_experimental_results}.

For each width $d \in\{3,\dots,20\}$, we draw $2048$ training examples and $512$ test examples. Each of the five seeds fixes the teacher, the training and test inputs, and the student initialization, i.e. the five seeds are five independent teacher-student problems.  Students are trained with full-batch Adam for $4000$ optimization steps using learning rate $10^{-3}$ and  $\beta_1 = 0.9, \beta_2 = 0.999, \varepsilon = 10^{-8}$ and no weight decay. The task loss is mean-squared error to the teacher outputs. For the projection experiments, we use the reference point $\theta_{\text{ref}}=0$. We ablate this choice in Fig.~\ref{fig:ablation}. 

\subsubsection{MNIST}\label{app: data: mnist}
For the classification experiment, we train width-scaled MLPs on MNIST. Images are flattened to vectors in $\mathbb R^{784}$, normalized using the empirical mean and standard deviation of the training set.

The model is a two-hidden-layer ReLU MLP
\[
784 \to h_1 \to h_2 \to 10.
\]
In width-scaled runs, we set $h_1=h_2=w$ and vary the width 
$w\in\{32, 64, 96, 128, 160\}$. 
The respective parameter counts are $26{,}506$, $55{,}050$, $85{,}642$, $118{,}282$ and $152{,}970$.

We train with Adam for $2000$ optimization steps using learning rate $10^{-3}$, $\beta_1 = 0.9, \beta_2 = 0.999, {\varepsilon = 10^{-8}}$, mini-batch size $64$, no weight decay and cross-entropy loss. All MNIST results are obtained from a single training run per width.
The final test accuracies are $95.3\%$, $96.7\%$, $97.0\%$, $97.1\%$ and $96.9\%$ for $w = 32, 64, 96, 128$ and $160$.   Since we compute projection curves only for the leading $12{,}000$ eigendirections, we stay slightly below these accuracies, which motivates the use of thresholds up to $\tau=0.92$. This is close to the accuracy of a linear classifier on MNIST.
  For the projection experiments, we use the initialization as reference point $\theta_{\text{ref}}=\theta_0$ and $10{,}000$ training images as probe set. Here $f_\theta(x)\in\mathbb R^{10}$ denotes the logits, so the behavioral loss and the diagonal of $H_{\mathrm{beh}}$ used for OBD are computed on the pre-softmax outputs.

\subsection{Implementation}
\label{app:implementation}
All experiments are implemented in PyTorch and were run on NVIDIA RTX 2080 Ti, TITAN RTX or RTX A6000. In total, the reported MNIST experiments required approximately $150$ GPU-hours of compute; we note that this is an estimate given the heterogeneous compute environment. The code to reproduce all experiments is available at \url{https://github.com/Maascha/functional-degeneracy}.

Since explicitly forming the behavioral Hessian is computationally expensive, we instead access it through matrix-free eigensolves. For this, we work with the equivalent dual operator $|P|^{-1}J_PJ_P^\top$ in probe-output space (see App.~\ref{app:behavioral_hessian}), implemented via Jacobian-vector and vector-Jacobian products. 
We estimate the leading eigenspaces using symmetric Lanczos iteration with full orthogonalization~\cite{saad2011numerical}. In the teacher-student experiments, the number of Lanczos steps is chosen as $\min(D,\max(4k,60))$ for a target of $k$ leading eigenpairs, where $D=|\mathcal P|m$ is the dimension of the operator. Eigenpairs are retained by the scale-invariant criterion $\lambda_j \geq 10^{-6}\lambda_1$, since an absolute cutoff would depend on the output scale, which varies across widths. Projection curves use only these retained eigendirections. If a tolerance is not reached within them, the rank is undefined for that seed, and we exclude the seed in the figures. Parameter-space eigenvectors are recovered lazily as $v_j = J_{\mathcal{P}}^\top u_j / \|J_{\mathcal{P}}^\top u_j\|$ and, for the teacher-student experiments, re-orthonormalized by modified Gram-Schmidt before being accumulated.

For MNIST, the operator has dimension $D=10^5$, and a single run for the $12{,}000$ leading eigenpairs exceeds the available GPU memory. We therefore compute the eigenpairs in chunks of $1000$. Each chunk is a new Lanczos run of $4000$ steps with full orthogonalization, applied to the operator deflated by the eigenvectors of the previous chunks. The new eigenvectors are orthogonalized against the previous ones, and their eigenvalues are computed as Rayleigh quotients of the original operator. The recovered parameter-space eigenvectors are used without further re-orthonormalization. At every width, all $12{,}000$ eigenvalues exceed $10^{-6}\lambda_1$.

\section{Ordering of the eigendirections}\label{app:recovery_variants}

Throughout the main paper $\hat r_q$ and $r_\tau$ project onto the
top-$k$ eigendirections ordered by decreasing eigenvalue $\lambda_j$.

We will derive a different version, which yields a smaller rank in all our experiments. For this, let $\{v_j\}$ be the orthonormal eigenvectors of $H_{\mathrm{beh}}(\theta_\ast)$ with eigenvalues $\lambda_j$, and expand the learned displacement
$\theta_\ast-\theta_{\mathrm{ref}}$ in this basis.  Retaining an index set $S$ gives the reconstruction
\[
\hat\theta_S
=\theta_{\mathrm{ref}}
+\sum_{j\in S}\langle v_j,\theta_\ast-\theta_{\mathrm{ref}}\rangle\, v_j ,
\]
so the unrecovered component of the displacement is
\[
\hat\theta_S-\theta_\ast
=-\sum_{j\notin S}\langle v_j,\theta_\ast-\theta_{\mathrm{ref}}\rangle\,v_j ,
\qquad
\big\|\hat\theta_S-\theta_\ast\big\|^2
=\sum_{j\notin S}\langle v_j,\theta_\ast-\theta_{\mathrm{ref}}\rangle^2 .
\]
Evaluating the second-order model of the behavioral loss at $\hat\theta_S$ and
using orthonormality of the eigenbasis gives the exact identity
\begin{equation}
\frac12\big(\hat\theta_S-\theta_\ast\big)^\top H_{\mathrm{beh}}(\theta_\ast)
\big(\hat\theta_S-\theta_\ast\big)
=\frac12\sum_{j\notin S}
\lambda_j\langle v_j,\theta_\ast-\theta_{\mathrm{ref}}\rangle^2 ,
\label{eq:residual_separable}
\end{equation}
and $L_{\mathrm{beh}}(\hat\theta_S;\theta_\ast,\mathcal P)$ agrees with
\eqref{eq:residual_separable} up to $o(\|\hat\theta_S-\theta_\ast\|^2)$. It is minimized when the eigendirections are ordered decreasing according to the corresponding  $\lambda_j\langle v_j,\theta_\ast-\theta_{\mathrm{ref}}\rangle^2$. We define the corresponding statistic as the \textit{loss-optimal behavioral recovery rank}  $r_\tau^{\mathrm{opt}}$.

For two reasons,  the eigenvalue ordering is our main definition. First, $\{\lambda_j,v_j\}$ depend only on
$H_{\mathrm{beh}}(\theta_\ast)$, and hence on the parameter vector $\theta_\ast$ and the probe distribution, whereas $\lambda_j\langle v_j,\theta_\ast-\theta_{\text{ref}}\rangle^2$ depends in addition on the displacement $\theta_\ast-\theta_{\mathrm{ref}}$. 
As a result, the loss-optimal rank mixes
local geometry with the training trajectory. Second, the loss-optimal ordering 
can only be applied within the whole eigenspace that has actually been computed, with the other definition only requiring the top of the spectrum.

\section{Behavioral Hessian geometry}
\label{app:behavioral_hessian}
This appendix develops the geometric interpretation of the behavioral Hessian introduced in Sec.~\ref{sec:methodology} and relates it to several established objects in optimization and representation theory.
The sections are tied together through a common geometric object: the Jacobian of the finite-probe evaluation map, $J_{\mathcal P}(\theta_\ast)$. Each section studies how different notions of local model geometry arise from this same Jacobian, but viewed from different perspectives and in different spaces.
\begin{itemize}
    \item In Sec.~\ref{app: sub behavioural}, the Jacobian defines the behavioral Hessian as the parameter-space Gram matrix $J_{\mathcal P}^\top J_{\mathcal P}$, which characterizes the local sensitivity of model outputs to parameter perturbations. 
    
    \item Sec.~\ref{app:behavioral_hessian_task} shows that the same structure appears within the generalized Gauss-Newton component of the task-loss Hessian, thereby connecting behavioral geometry to standard optimization curvature.
    \item In Sec.~\ref{app:behavioral_hessian_fd}, the rank and kernel of the Jacobian are related to functional dimension, linking flat parameter directions to infinitesimal function-preserving transformations. 
    \item Sec.~\ref{app:behavioral_hessian_ntk} then considers the dual object $J_{\mathcal P}J_{\mathcal P}^\top$, namely the empirical NTK, which acts in function space rather than parameter space while sharing the same nonzero spectrum.
    \item Finally, Sec.~\ref{app:IG} interprets the same Gram structure statistically as an empirical Fisher information matrix, providing an information-geometric interpretation in terms of local identifiability and informative directions.
\end{itemize}
Taken together, the appendix shows that behavioral curvature, optimization curvature, functional degeneracy, NTKs, and Fisher information can all be understood as different manifestations of the same local Jacobian geometry induced by the probe evaluation map.

\subsection{Behavioral Hessian geometry}\label{app: sub behavioural}
 In this section, we discuss the behavioral Hessian introduced in Sec.~\ref{sec:methodology}. Throughout this appendix, let $\theta_\ast\in\mathbb{R}^p$ be a
trained checkpoint, $m = \operatorname{dim}(f_\theta(x))$ the output dimension, and let $\mathcal P$ be a finite probe set. 
We define the finite-probe evaluation map
\[
F_{\mathcal P}(\theta)
=
\bigl(f_\theta(x)\bigr)_{x\in\mathcal P}
\in \mathbb R^{|\mathcal P|m},
\]
where $m$ is the output dimension and we stack all outputs. Let
\[
J_{\mathcal P}(\theta)
=
\frac{\partial F_{\mathcal P}(\theta)}{\partial \theta}
\]
denote its Jacobian.
We can now write the behavioral loss in vector notation as 
\[
L_{\mathrm{beh}}(\theta;\theta_\ast,\mathcal P)
=
\frac{1}{2|\mathcal P|}
\left\|
F_{\mathcal P}(\theta)
-
F_{\mathcal P}(\theta_\ast)
\right\|_2^2 .
\]
Differentiating once gives
\[
\nabla_\theta L_{\mathrm{beh}}(\theta;\theta_\ast,\mathcal P)
=
\frac{1}{|\mathcal P|}
J_{\mathcal P}(\theta)^\top
\left(
F_{\mathcal P}(\theta)
-
F_{\mathcal P}(\theta_\ast)
\right).
\]
Differentiating again yields
\[
\nabla_\theta^2 L_{\mathrm{beh}}(\theta;\theta_\ast,\mathcal P)
=
\frac{1}{|\mathcal P|}
J_{\mathcal P}(\theta)^\top
J_{\mathcal P}(\theta)
+
\frac{1}{|\mathcal P|}
\sum_{i=1}^{|\mathcal P|m}
\left(
F_{\mathcal P}(\theta)
-
F_{\mathcal P}(\theta_\ast)
\right)_i
\nabla_\theta^2
\bigl(F_{\mathcal P}(\theta)\bigr)_i.
\]
At $\theta=\theta_\ast$, the residual is zero. Therefore the second term vanishes, and the behavioral Hessian is the Jacobian Gram matrix
\begin{equation}
H_{\mathrm{beh}}(\theta_\ast;\mathcal P)
= \nabla_\theta^2 L_{\mathrm{beh}}(\theta_\ast;\theta_\ast,\mathcal P) = \frac{1}{|\mathcal P|} J_{\mathcal P}(\theta_\ast)^\top J_{\mathcal P}(\theta_\ast). \label{eq:beh_hessian_jacobian}
\end{equation}
We also highlight that by Equation \eqref{eq:beh_hessian_jacobian},
\[
\operatorname{rank}\bigl(H_{\mathrm{beh}}(\theta_\ast;\mathcal P)\bigr)
=
\operatorname{rank}\bigl(J_{\mathcal P}(\theta_\ast)\bigr)
\le
\min\{p,|\mathcal P|m\}.
\]
Thus the behavioral Hessian is necessarily rank-limited by both parameter dimension and probe-output dimension. In particular, if $|\mathcal P|m<p$, then its nullspace has dimension at least $p-|\mathcal P|m$.

\subsection{Connection to the task loss Hessian}
\label{app:behavioral_hessian_task}
Let $\mathcal X=\{x_i\}_{i=1}^N$ denote the training data, $\{y_i\}_{i=1}^N$ corresponding outputs and consider the supervised task loss
\[
L_{\mathrm{task}}(\theta)
=
\frac{1}{N}
\sum_{i=1}^N
\ell(f_\theta(x_i),y_i).
\]
Define the
corresponding stacked evaluation map
$
F_{\mathcal X}(\theta)
=
\bigl(f_\theta(x_i)\bigr)_{i=1}^N
\in \mathbb R^{Nm}$ with
$J_{\mathcal X}(\theta) = \frac{\partial F_{\mathcal X}(\theta)}{\partial \theta}$ its Jacobian.

The Hessian of the task loss at $\theta_\ast$ decomposes into a
generalized Gauss-Newton term and a curvature-residual term:
\[
\nabla_\theta^2 L_{\mathrm{task}}(\theta_\ast)
=
H_{\mathrm{GN}}(\theta_\ast)
+
R_{\mathrm{curv}}(\theta_\ast)
\]
with
\[
H_{\mathrm{GN}}(\theta_\ast) = J_{\mathcal X}(\theta_\ast)^\top
\left[
\frac{1}{N} \operatorname{blockdiag}_{i=1}^N
\left(
\nabla_1^2 \ell(a,y_i) \big|_{a=f_{\theta_\ast}(x_i)}
\right)
\right]
J_{\mathcal X}(\theta_\ast),
\]
where the Hessian $\nabla_1^2\ell$ is taken with respect to the first component, the model output $f_\theta(x_i)$.
The remaining term is
\[
R_{\mathrm{curv}}(\theta_\ast) = \frac{1}{N} \sum_{i=1}^N \sum_{j=1}^m \frac{\partial \ell}{\partial f_j} \bigl(f_{\theta_\ast}(x_i),y_i\bigr)  \nabla_\theta^2 f_{\theta,j}(x_i) \big|_{\theta=\theta_\ast}.
\]
If the probe set is chosen as the training set, $\mathcal{P} = \mathcal{X}$, and $\ell(a,y)=\tfrac12\|a-y\|^2$ is the squared error, the behavioral Hessian becomes the Gauss-Newton component of the task Hessian:
\[
H_{\mathrm{beh}}(\theta_\ast;\mathcal P)
=
H_{\mathrm{GN}}(\theta_\ast).
\]

\subsection{Connection to functional dimension}
\label{app:behavioral_hessian_fd}

Following \citet{grigsby2025functional}, the batch functional dimension at
$\theta_\ast$ on a finite probe set $\mathcal P$ is
\[
\operatorname{FD}_{\mathcal P}(\theta_\ast)
=
\operatorname{rank}
\bigl(J_{\mathcal P}(\theta_\ast)\bigr).
\]
The full local functional dimension is obtained by taking the supremum over
finite probe sets,
\[
\operatorname{FD}(\theta_\ast)
=
\sup_{\mathcal Q}
\operatorname{rank}
\bigl(J_{\mathcal Q}(\theta_\ast)\bigr).
\]
Thus $\operatorname{FD}_{\mathcal P}(\theta_\ast)\leq
\operatorname{FD}(\theta_\ast)$.

We know from Equation \eqref{eq:beh_hessian_jacobian} that the behavioral Hessian is the Jacobian Gram matrix $J_{\mathcal P}^\top J_{\mathcal P}$, which has the same kernel as $J_{\mathcal P}$ and hence
\[
\operatorname{rank}
\bigl(H_{\mathrm{beh}}(\theta_\ast;\mathcal P)\bigr)
=
\operatorname{FD}_{\mathcal P}(\theta_\ast)
\leq
\operatorname{FD}(\theta_\ast).
\]

This connects our local loss-geometry statistic to exact functional
degeneracy: Directions in the kernel for all finite probe sets are infinitesimal degeneracies of the realization map: to first order, they do not change the represented function.

\paragraph{Remark (reparameterization).}
Consider a hidden unit of a ReLU network with incoming weights $w_{\mathrm{in}}$, bias $b$ and outgoing weights $w_{\mathrm{out}}$. Replacing these parameters by $(c\,w_{\mathrm{in}}, c\,b, w_{\mathrm{out}}/c)$ leaves $f_\theta$ unchanged for every $c > 0$. We write such a rescaling of one or several units as $\theta \mapsto D\theta$ with a positive diagonal matrix $D$. Since $F_{\mathcal{P}}(D\theta) = F_{\mathcal{P}}(\theta)$, the chain rule gives
\begin{equation*}
J_{\mathcal{P}}(D\theta) = J_{\mathcal{P}}(\theta)\, D^{-1},
\qquad
H_{\mathrm{beh}}(D\theta_\ast; \mathcal{P}) = D^{-1} H_{\mathrm{beh}}(\theta_\ast; \mathcal{P})\, D^{-1}.
\end{equation*}
The rank of $H_{\mathrm{beh}}$, and hence $\mathrm{FD}_{\mathcal{P}}$, is invariant under rescaling, while its eigenvalues, its eigenvectors and the projection $\Pi_k$ change in general. This result can also be observed by considering the relation between the FIM and the Hessian together with the fact that the FIM is covariant under appropriate reparameterization of a statistical model \cite{calin2014geometric}. The behavioral recovery rank therefore depends on the parameter vector $\theta_\ast$ and can differ between parameter vectors that realize the same function. 

The behavioral Hessian accounts for the rescaling symmetry to first order. For a single hidden unit, the rescaled parameters form a curve in $c$ that passes through $\theta_\ast$ at $c = 1$. The derivative of this curve at $c = 1$ is the direction $t = (w_{\mathrm{in}}, b, -w_{\mathrm{out}})$, with zeros in all other coordinates. Since the outputs on the probe set are constant along the curve, we obtain $J_{\mathcal{P}}(\theta_\ast)\, t = 0$ wherever $f$ is differentiable in $\theta$, and hence $H_{\mathrm{beh}}(\theta_\ast; \mathcal{P})\, t = 0$. Each hidden unit thus contributes one direction to the kernel of $H_{\mathrm{beh}}$. The retained eigenvectors have positive eigenvalues and are orthogonal to this kernel, so the reconstructed displacement $\hat{\theta}_k - \theta_{\mathrm{ref}}$ has no component along these directions.

\subsection{Connection to the empirical NTK}
\label{app:behavioral_hessian_ntk}

The same evaluation-map Jacobian also defines the empirical neural tangent
kernel on the probe set. The empirical NTK at
$\theta_\ast$ is
\[
K_{\mathcal P}(\theta_\ast)
= \frac{1}{|\mathcal P|}
J_{\mathcal P}(\theta_\ast)
J_{\mathcal P}(\theta_\ast)^\top
\in
\mathbb R^{|\mathcal P|m\times |\mathcal P|m}.
\]

Thus the behavioral Hessian and the empirical NTK have the same nonzero eigenvalues as well as the same rank. The two matrices act on different spaces. The empirical NTK acts on probe-set output perturbations, whereas the behavioral Hessian acts on \emph{parameter} perturbations. 

\subsection{Connection to Information Geometry } \label{app:IG}

The identification of the behavioral Hessian as a Jacobian Gram matrix in Equation \eqref{eq:beh_hessian_jacobian} gives us a natural connection to information geometry (IG) \cite{Amari2000information}. 
As a central element in IG, we study the  Fisher Information Matrix (FIM) which is defined as the expected outer product of the log-likelihood gradients of a parameterized distribution  $p(y|x, \theta)$:
\[
\mathcal{I}(\theta) = \mathbb{E}_{x \sim \mathcal P,  y\sim p(\cdot|x,\theta)} \left[ \nabla_\theta \log p(y|x, \theta) \nabla_\theta \log p(y|x, \theta)^\top \right];
\]
which can equivalently be expressed as the negative expectation of the Hessian.

In the following we outline how this can be connected to the behavioral loss. Starting from the stacked evaluation map
$
F_{\mathcal P}(\theta)
=
\bigl(f_\theta(x)\bigr)_{x\in\mathcal P}\in \mathbb R^{|\mathcal P|m}$,
the loss
\[
L_{\mathrm{beh}}(\theta) = \frac{1}{2|\mathcal{P}|}\left\|F_{\mathcal{P}}(\theta)- F_{\mathcal{P}}(\theta_\ast) \right\|^2
\]
can be reinterpreted as a \emph{negative log-likelihood}\footnote{Remark: this is not a unique choice for building a statistical model but illustrates how the connection to the behavioral loss can be established formally. This model is not meant to describe a real data-generating process globally, instead it captures local deviations in the function space and instead should be viewed as a local statistical model that probes how parameters affect outputs near $\theta_\ast$.} under a Gaussian observation model on the outputs
\[
F_{\mathcal{P}}(\theta_\ast) = F_{\mathcal{P}}(\theta) + \xi,\quad\xi\sim\mathcal{N}(0,I)
\]
equivalently we can write
\[
p(F_{\mathcal{P}}(\theta_\ast)~|~\theta) \propto\exp\left(-|\mathcal{P}|L_{\mathrm{beh}}(\theta)\right)
\]
where locally around $\theta_\ast$, we have defined a parametric statistical model over the probe set output and
\[
L_{\mathrm{beh}}(\theta) =- \frac{1}{|\mathcal{P}|}\log p(F_\mathcal{P}(\theta_\ast)~|~\theta)+\mathrm{constant}
\]
Therefore the FIM naturally emerges as 
\begin{align*}
    \mathcal{I}(\theta) 
    &= \mathbb{E}_{x \sim \mathcal P, y\sim p(\cdot|x,\theta)} \left[ \nabla_\theta \log p(y|x, \theta) \nabla_\theta \log p(y|x, \theta)^\top \right]\\
    &=\frac{1}{|\mathcal{P}|}J_\mathcal{P}(\theta)^\top J_\mathcal{P}(\theta).
\end{align*}
So at $\theta_\ast$,
\[
\mathcal{I}(\theta_\ast) = H_{\mathrm{beh}}(\theta_\ast;\mathcal{P}),
\]
and the behavioral Hessian is exactly the Fisher Information Matrix of this probe-induced model.

This gives us directly the interpretation that behavioral degeneracy is linked to  statistical non-identifiability under the probe distribution, which directly links back to the previously discussed notion of functional dimension.
Similarly, the rank can be interpreted as an information-geometric notion where the behavioral recovery rank $\hat r_q$ becomes the minimal number of informationally significant directions needed to recover performance.
Finally, we can recall that we previously noted
\[
H_{\mathrm{beh}} = \frac{1}{|\mathcal{P}|}J^\top J,\quad K_\mathcal{P} = \frac{1}{|\mathcal{P}|}JJ^\top.
\]
From an IG viewpoint, we can therefore see $H_{\mathrm{beh}}$ as a metric on the parameter space and $K_\mathcal{P}$ as a metric on the function space, which describes a dual relationship via the evaluation map. 

\section{Pruning with refitting}
\label{app:pruning_refit}

The baselines in Sec.~\ref{sec:methodology} delete parameters and leave the remainder untouched. We additionally report a \emph{refitted} variant in which
the output layer is refitted after each removal. Since the network output is linear in the final layer given the surviving hidden activations, the optimal weights for a fixed set of retained units are available in closed form as a least-squares solution, so no gradient steps are required and the procedure remains one-shot. The refit is computed on the training split and respects the sparsity pattern, i.e. entries removed by pruning are held at zero and only the surviving support is re-solved.

The refitted variant quantifies  whether the retained
units still span enough of the function, since their contributions are recombined optimally afterwards. This is a substantial intervention, since the output layer comprises roughly half of the weights in these two-layer students.

\begin{figure}[t]
    \centering

    \subfigure[Noiseless.]{
    \includegraphics[width=.38\textwidth]{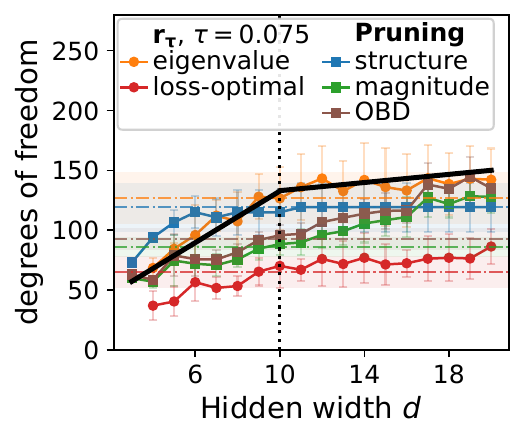}
    }
    \hspace{.8cm}
    \subfigure[$\sigma^2=0.01$]{    \includegraphics[width=.38\textwidth]{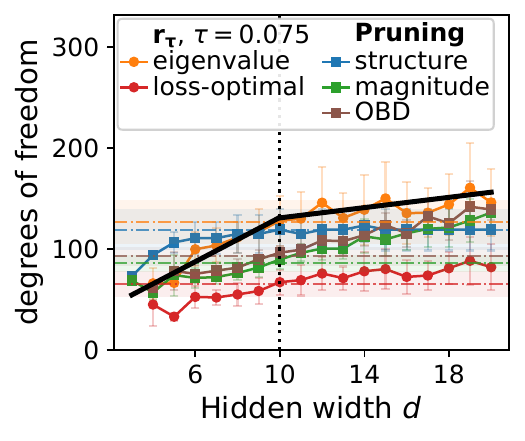}}
    \caption{We repeat the pruning experiments of Fig.~\ref{fig:2regimes}(c), refitting the student's output layer in closed form after each removal. The dash-dotted lines show a teacher control: the same pipeline applied to the teacher network itself. With refitting, structure pruning curves flatten similarly to the recovery rank, approaching the teacher's compressible level. Magnitude pruning and OBD still increase notably.}
    \label{fig:pruning_refit}
\end{figure}

Fig.~\ref{fig:pruning_refit} repeats Fig.~\ref{fig:2regimes}(c) under this refitting, together with the same quantities measured on the teacher network itself (dash-dotted). The retained number of parameters falls substantially for structure pruning, and greedy neuron removal no longer grows with student width. Part of the gap reported in Sec.~\ref{sec: results} therefore reflects the absence of compensation, not the coordinate basis alone, although all three baselines remain above the loss-optimal recovery rank. We do not repeat this variant on MNIST, since the closed form relies on a squared-error objective and has no counterpart under cross-entropy.

\section{Further experimental results}\label{app:further_experimental_results}
In this appendix, we provide further experimental results that were omitted in the main paper for brevity.

\subsection{Teacher-student Experiments}

\begin{figure}[t]
    \centering
    \subfigure[Test MSE.]{ 
    \includegraphics[width=.48\textwidth]{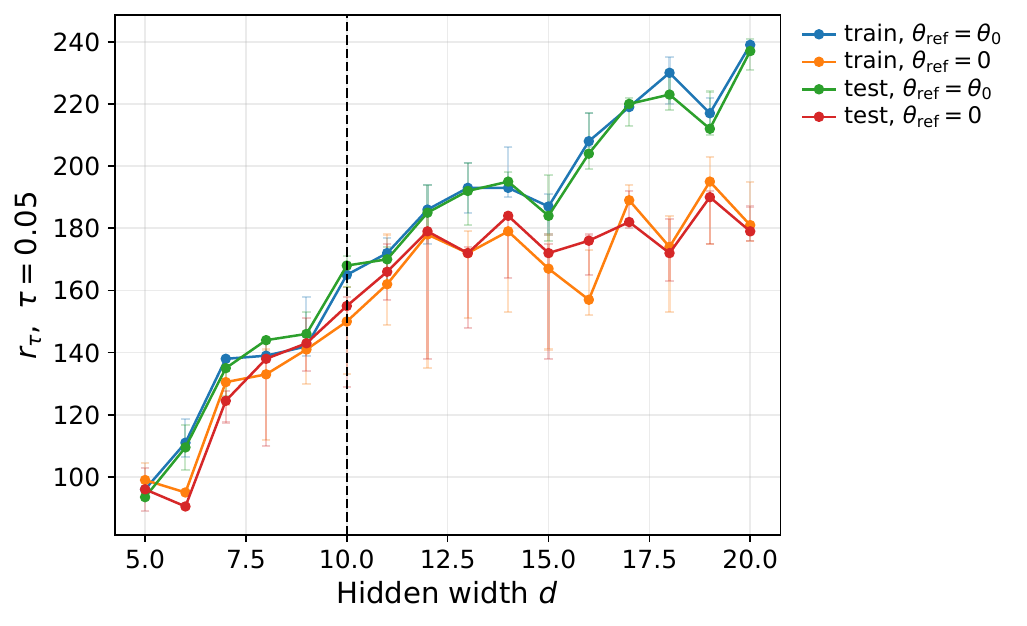}}
    \hfill
     \subfigure[Train MSE.]{ 
    \includegraphics[width=.48\textwidth]{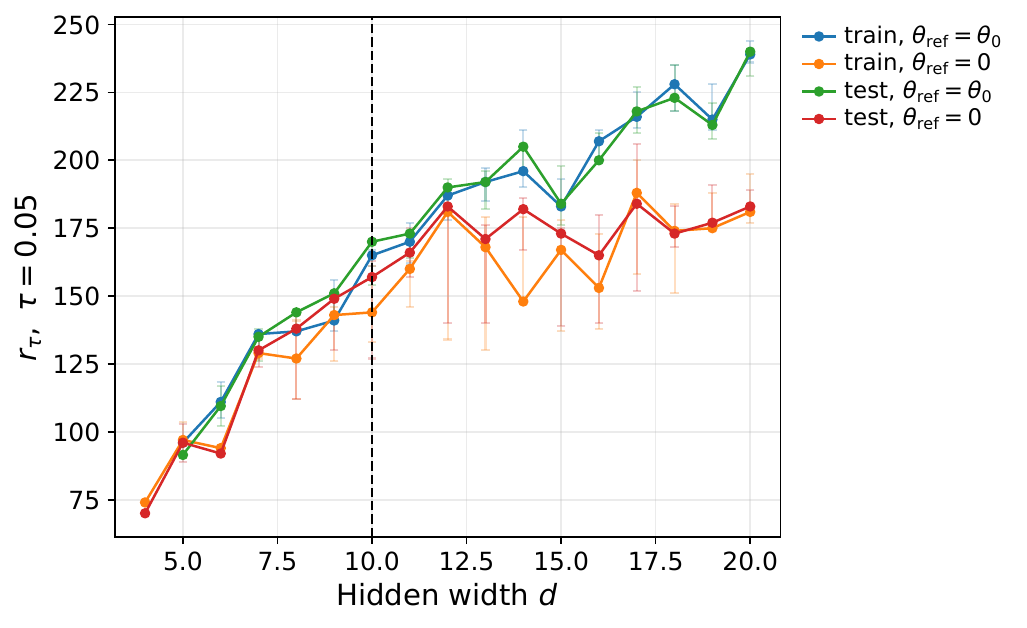}}
    \caption{Ablation experiments, confirming robustness under reconstruction choices. We vary the probe set and  $\theta_{\text{ref}}$ for $\tau=0.05$, median and interquartile range over the seeds with a defined rank. The behavioral  recovery rank growth is larger in the task-saturated regime for $\theta_{\text{ref}}=\theta_0$ in comparison to $\theta_{\text{ref}}=0$, but still flattens more  than in the capacity-limited regime. }
    \label{fig:ablation}
\end{figure}

\begin{figure}[t]
    \centering
   \subfigure[Same as Fig.~\ref{fig:2regimes} (b), but the test set is used as probe set.]{\includegraphics[height=3.2cm,trim=2mm 1mm 41mm 1mm, clip]{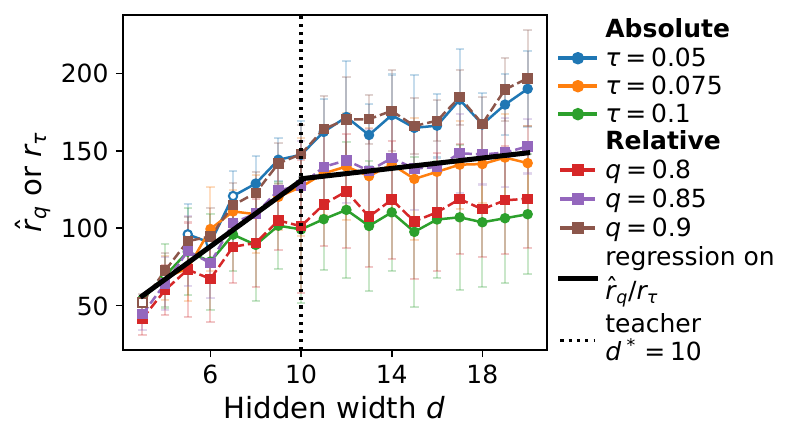}}
    \hfill
 \subfigure[Same as Fig.~\ref{fig:2regimes} (b), but $\theta_{\text{ref}}=\theta_0$.]{\includegraphics[height=3.2cm,trim=2mm 1mm 41mm 1mm, clip]{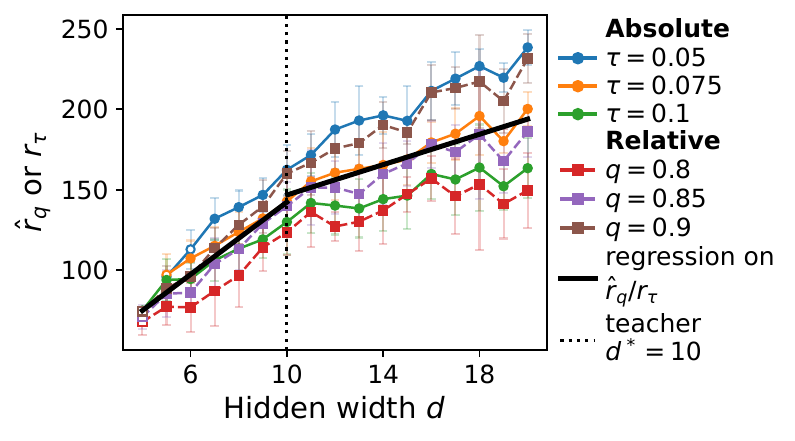}}
\hfill
 \subfigure[Same as Fig.~\ref{fig:2regimes} (b), but train MSE, not test MSE]{\includegraphics[height=3.2cm,trim=2mm 1mm 3mm 1mm, clip]{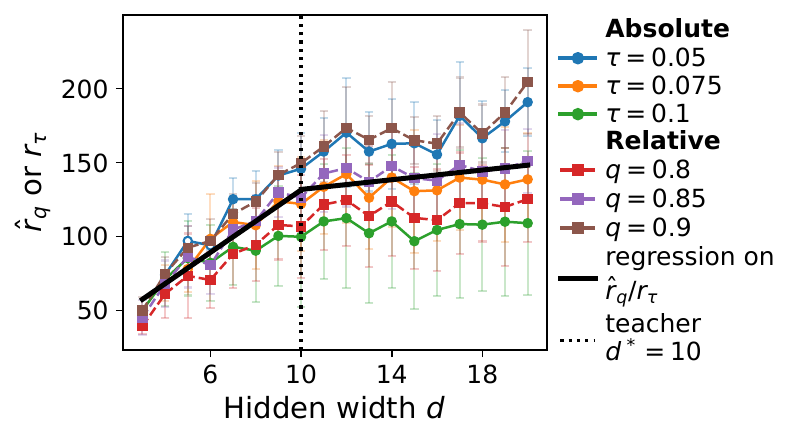}}
    \caption{Ablation experiments: Detailed curves for different behavioral rank thresholds, showing the ranks for selected curves of Fig.~\ref{fig:ablation}. }
    \label{fig:ablation2}
\end{figure}

\textbf{Changes of probe set and reference point.}
Figures~\ref{fig:ablation} and \ref{fig:ablation2} suggest that the recovery behavior is qualitatively stable under changes of probe set and reference point, suggesting that the observed trends are not tied to a particular reconstruction choice. For $\theta_{\text{ref}}=\theta_0$, the curve flattens less for widths exceeding teacher width, but still clearly.

\begin{figure}[t]
    \centering
    \subfigure[Test MSE for projecting on varying number of eigendirections.]{
    \includegraphics[height=3.2cm,trim=2mm 1mm 3mm 1mm, clip]{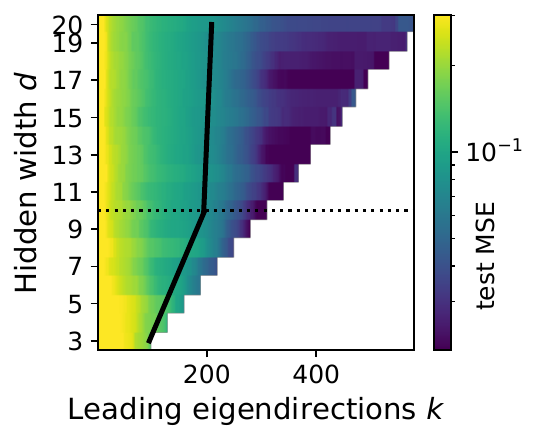}
    }
    \hfill
    \subfigure[Behavioral recovery rank extracted from the projection curves, mean and std. over the $5$ seeds. The hollow marker indicates that some seeds did not reach the threshold.]{
       \includegraphics[height=3.2cm,trim=2.5mm 1mm 3.5mm 1mm, clip]{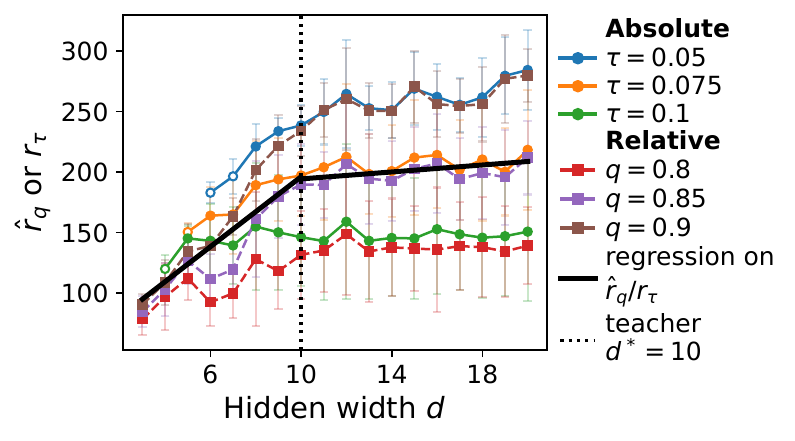} 
      }
      \hfill
          \subfigure[Pruning and behavioral recovery rank to loss threshold.]{
    \includegraphics[height=3.2cm,trim=2mm 1mm 2mm 1mm, clip]{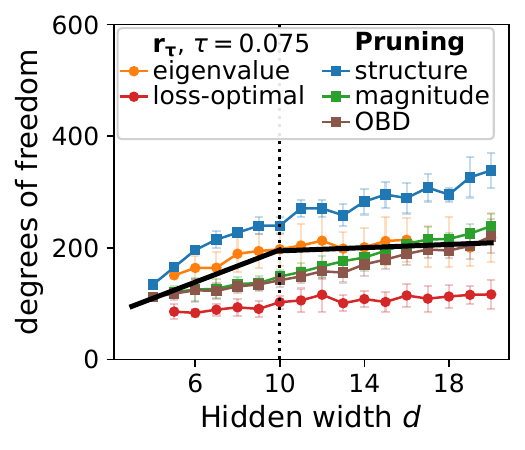}
    }
     \caption{Same setup as Fig.~\ref{fig:2regimes}, but with input dimension set to $20$.}
     \label{fig:ts_20_10}
\end{figure}

\begin{figure}[t]
    \centering
    \subfigure[Test MSE for projecting on varying number of eigendirections.]{
    \includegraphics[height=3.2cm,trim=2mm 1mm 3mm 1mm, clip]{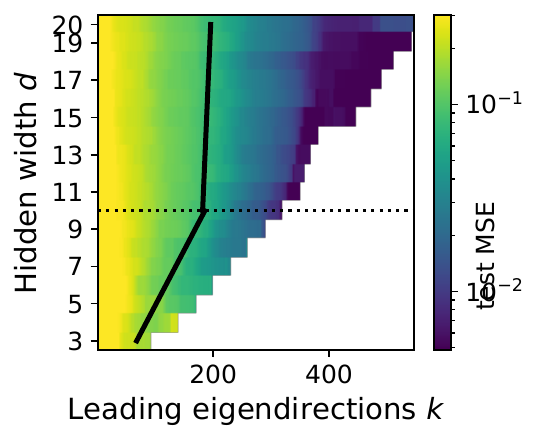}
    }
    \hfill
    \subfigure[Behavioral recovery rank extracted from the projection curves, mean and std. over up to $5$ runs.]{
       \includegraphics[height=3.2cm,trim=2.5mm 1mm 3.5mm 1mm, clip]{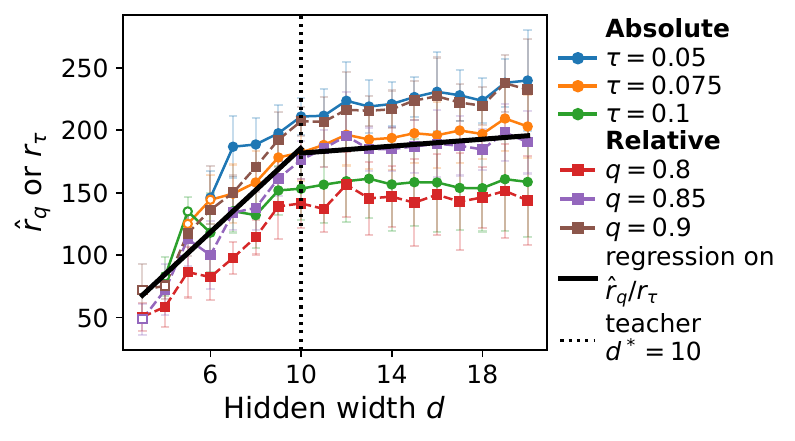} 
      }
      \hfill
          \subfigure[Pruning and behavioral recovery rank to loss threshold.]{
    \includegraphics[height=3.2cm,trim=2mm 1mm 2mm 1mm, clip]{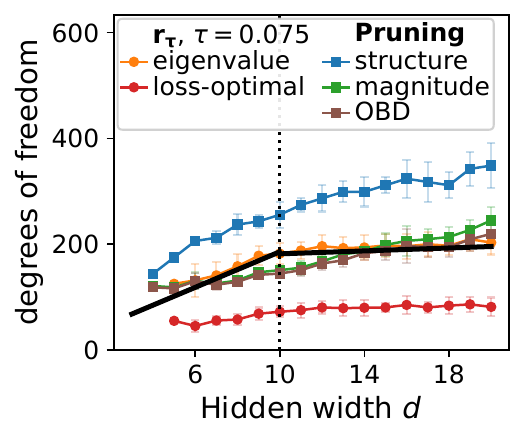}
    }
       \caption{Same setup as Fig.~\ref{fig:2regimes}, but with output dimension set to $20$.}
     \label{fig:ts_10_20}
\end{figure}

\begin{figure}[t]
    \centering
    \subfigure[Test MSE for projecting on varying number of eigendirections.]{
    \includegraphics[height=3.1cm,trim=2mm 1mm 3mm 1mm, clip]{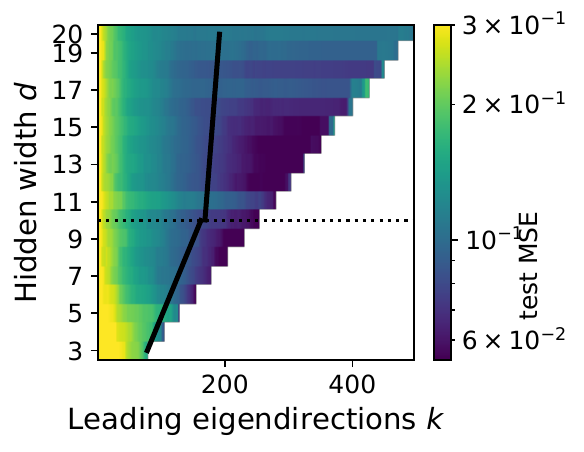}
    }
    \hfill
    \subfigure[Behavioral recovery rank extracted from the projection curves, mean and std. over up to $5$ runs.]{
       \includegraphics[height=3.1cm,trim=2.5mm 1mm 3.5mm 1mm, clip]{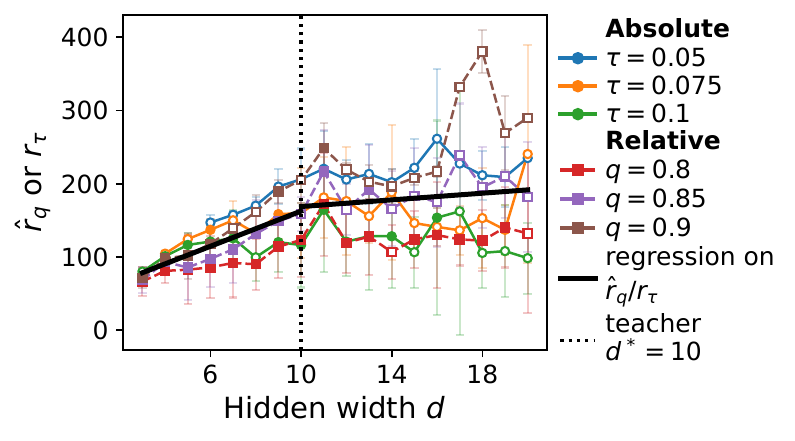} 
      }
      \hfill
          \subfigure[Pruning and behavioral recovery rank to loss threshold.]{
    \includegraphics[height=3.1cm,trim=2mm 1mm 2mm 1mm, clip]{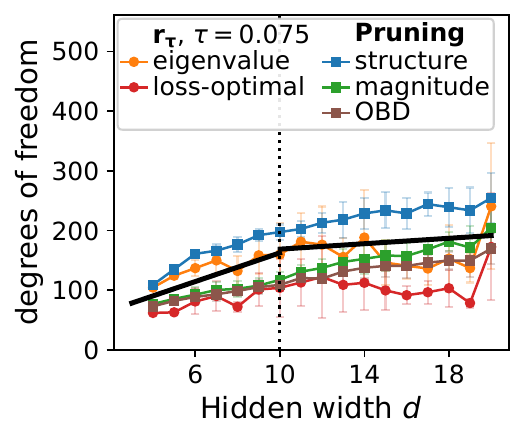}
    }
     \caption{Same setup as Fig.~\ref{fig:2regimes}, but with input dimension set to $20$, and output dimension $5$.}
     \label{fig:ts_20_5}
\end{figure}

\begin{figure}[t]
    \centering
    \subfigure[Test MSE for projecting on varying number of eigendirections.]{
    \includegraphics[height=3.2cm,trim=2mm 1mm 3mm 1mm, clip]{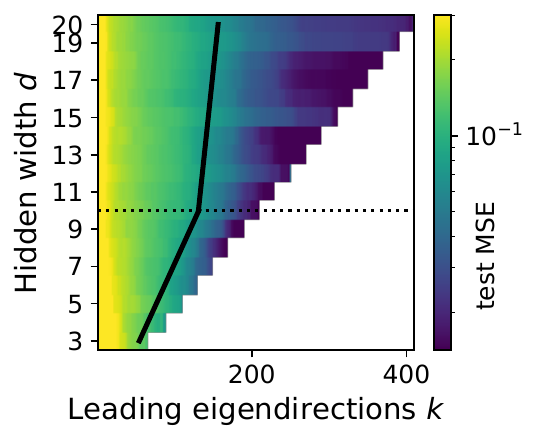}
    }
    \hfill
    \subfigure[Behavioral recovery rank extracted from the projection curves, mean and std. over up to $5$ runs.]{
       \includegraphics[height=3.2cm,trim=2.5mm 1mm 3.5mm 1mm, clip]{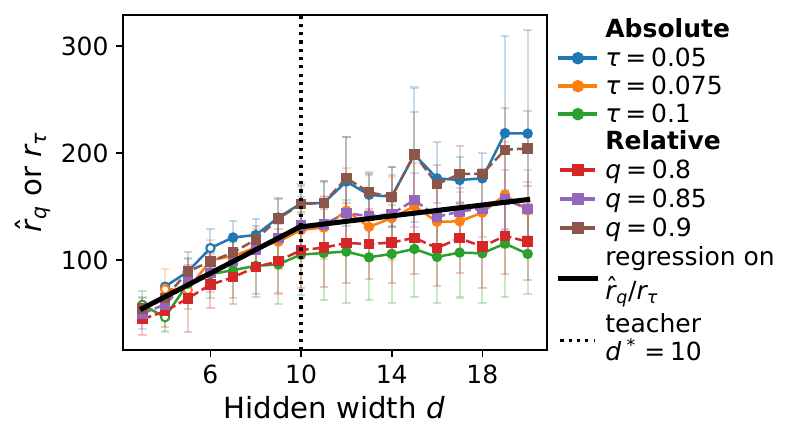} 
      }
      \hfill
          \subfigure[Pruning and behavioral recovery rank to loss threshold.]{
    \includegraphics[height=3.2cm,trim=2mm 1mm 2mm 1mm, clip]{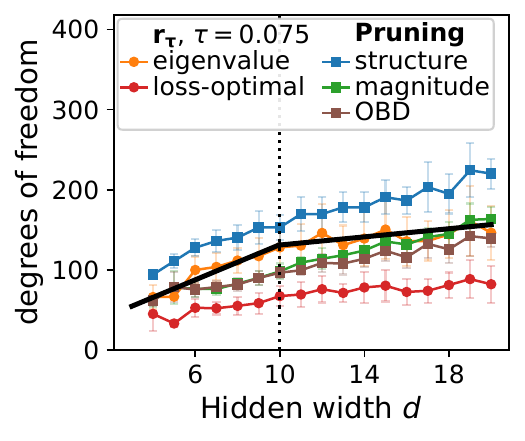}
    }
    \caption{Same setup as Fig.~\ref{fig:2regimes}, but with noise $\sigma^2=0.01$. The same regimes emerge.}
    \label{fig:ts_noise}
\end{figure}

\begin{figure}[t]
    \centering

    \subfigure[Noiseless.]{
    \includegraphics[width=.38\textwidth]{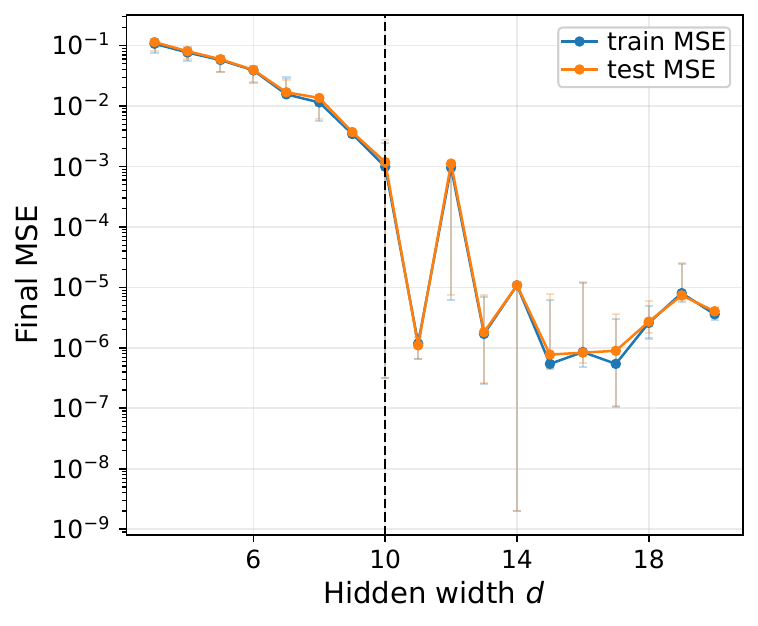}
    }
    \hspace{.8cm}
    \subfigure[$\sigma^2=0.01$]{    \includegraphics[width=.38\textwidth]{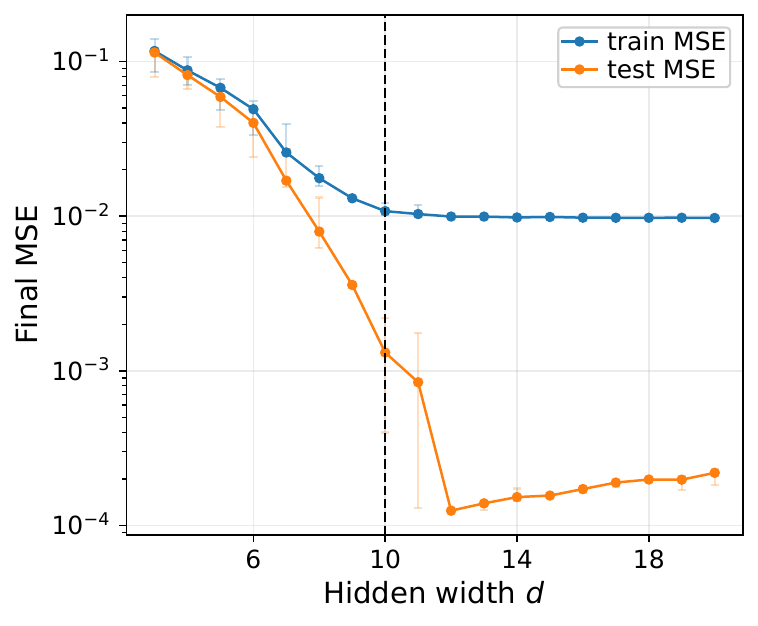}}
    \caption{We show final MSE for different student widths $d$. While in the noiseless setting, the training and test losses almost match, for the setting with noise, the students' train mean errors recover the noise level, with a slight double-descent-like behavior. The test data is noiseless. We show the median and the interquartile range over the five seeds.}
    \label{fig:ts_evol}
\end{figure}

\textbf{Changes of model architecture}
We change  parts of our teacher/student architecture to test the robustness of the findings and show Fig.~\ref{fig:2regimes} under these changes. In practice, we modify the input size (Fig.~\ref{fig:ts_20_10}), output size (Fig.~\ref{fig:ts_10_20}), and both (Fig.~\ref{fig:ts_20_5}). As an additional ablation, we add noise to the training data in our base architecture ($10 \to d \to 10$, described in App.~\ref{app: data student teacher}), shown in Fig.~\ref{fig:ts_noise}. Finally, Fig.~\ref{fig:ts_evol} compares the final MSE on the test/train set with and without noise, where a slight double descent is visible. %
These experiments suggest that our findings on the teacher-student model are robust across different  architectural parameters.

\begin{figure}[t]
    \centering
    \subfigure[Top eigenvalues of the behavioral Hessian, for different student widths.]{ 
    \includegraphics[width=.32\textwidth]{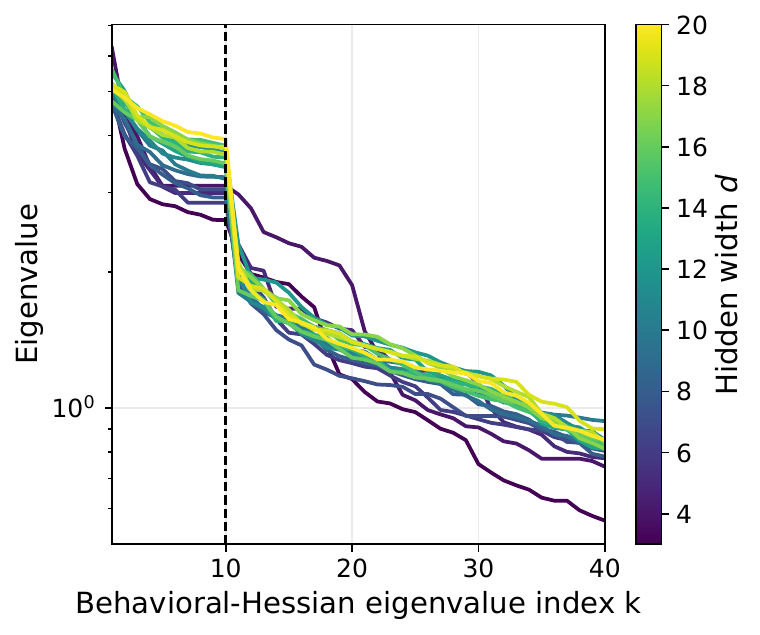}
    }
    \hfill
    \subfigure[Number of eigenvalues needed to reach $q \times \sum_i \lambda_i$. ]{    \includegraphics[width=.32\textwidth]{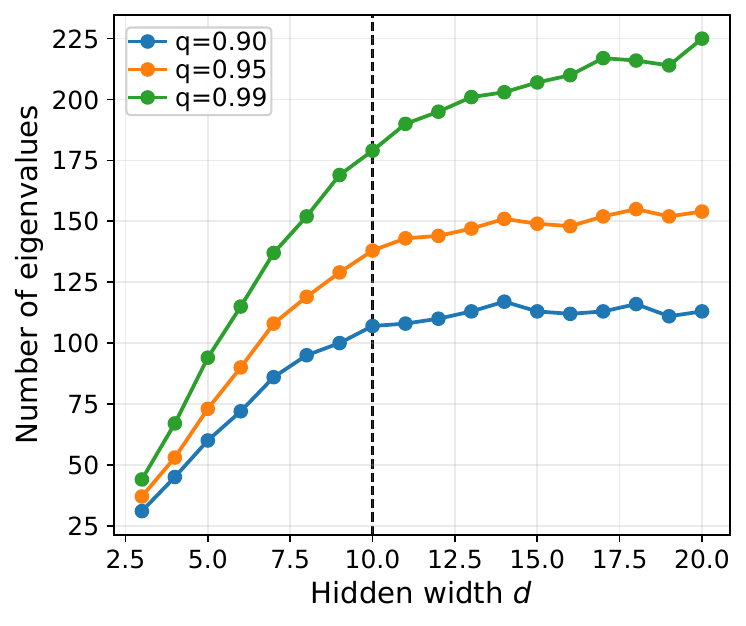}}
\hfill
     \subfigure[For different students, we plot the proportion of the top $k$ eigenvalues among all $ \sum_i^k \lambda_i/\sum_i \lambda_i$. Dotted lines indicate the thresholds $q=0.90, 0.95, 0.99$.]{  \includegraphics[width=.32\textwidth]{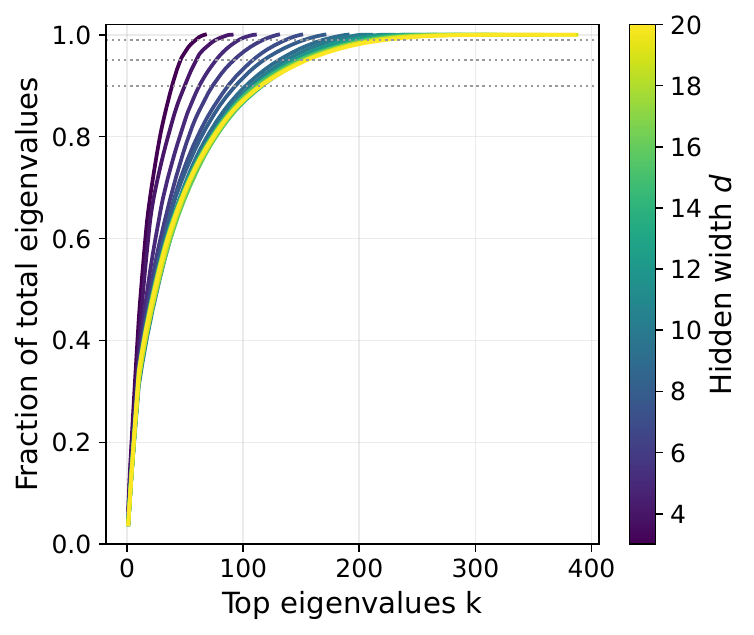}}
    \caption{Spectral quantities for the teacher-student experiments,  median over the five seeds.}
    \label{fig:eigenvalues_ts}
\end{figure}

\textbf{Spectral structure.}
The spectral structure of the behavioral Hessian further supports the interpretation of Fig.~\ref{fig:2regimes}. Fig.~\ref{fig:eigenvalues_ts}~(a) shows that the leading eigenvalues remain separated from a broad tail across widths, while Fig.~\ref{fig:eigenvalues_ts}~(b) and Fig.~\ref{fig:eigenvalues_ts}~(c) indicate that a limited subset of eigenmodes captures most of the spectral mass even as the parameter count grows.

\begin{figure}[h]
    \centering
    \subfigure[Noiseless.]{
    \includegraphics[width=.47\textwidth]{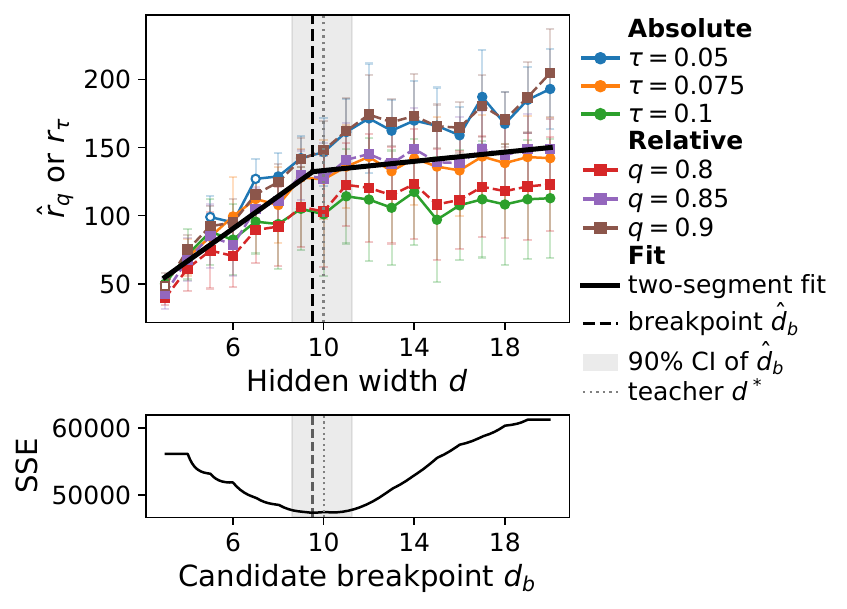}}
    \hfill
    \subfigure[$\sigma^2=0.01$.]{
    \includegraphics[width=.47\textwidth]{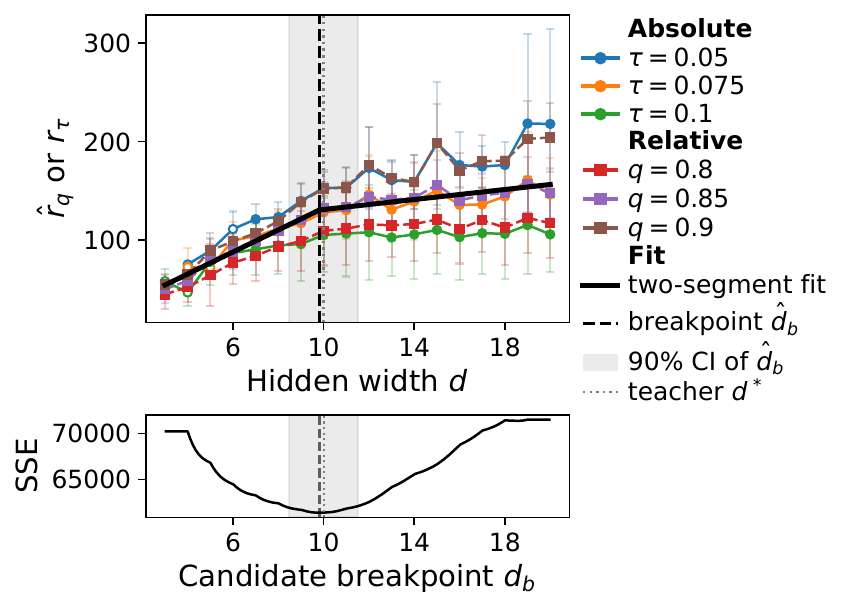}}
    \caption{Estimated breakpoint for the setups of Fig.~\ref{fig:2regimes}(b) and Fig.~\ref{fig:ts_noise}(b). Top: two-segment fit with the estimated breakpoint $\hat d_b$ and its $90\%$ bootstrap confidence interval. Bottom: SSE of the fit for each candidate breakpoint. Hollow markers indicate that fewer than five seeds reach the tolerance.}
    \label{fig:breakpoint}
\end{figure}

\textbf{Estimating the breakpoint.}
In Fig.~\ref{fig:2regimes}(b), the breakpoint of the regression is fixed at the teacher width $d^\ast=10$. In Fig.~\ref{fig:breakpoint}, we estimate it by fitting a continuous two-segment linear function for each candidate breakpoint and selecting the one with the smallest sum of squared errors (SSE). The estimate is $\hat d_b=9.5$ with a $90\%$ bootstrap confidence interval of $[8.6, 11.3]$, obtained by resampling the five seeds for each width and threshold $200$ times and repeating the breakpoint search on the resampled means. Similarly, for the setups of Figs.~\ref{fig:ts_20_10}--\ref{fig:ts_noise}, the estimates lie between $9.4$ and $10.9$. This supports that the growth of the behavioral recovery rank changes close to the teacher width also when the breakpoint is not imposed.

\begin{figure}[h]
    \centering
    \subfigure[Noiseless.]{
    \includegraphics[height=3.3cm]{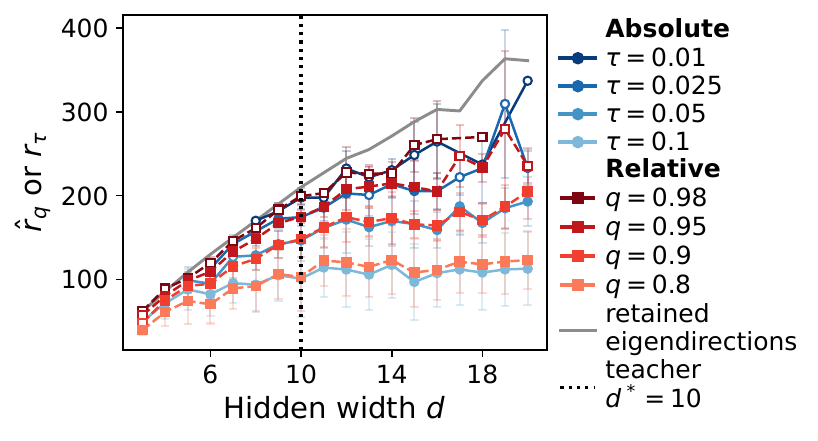}}
    \hfill
    \subfigure[$\sigma^2=0.01$.]{
    \includegraphics[height=3.3cm]{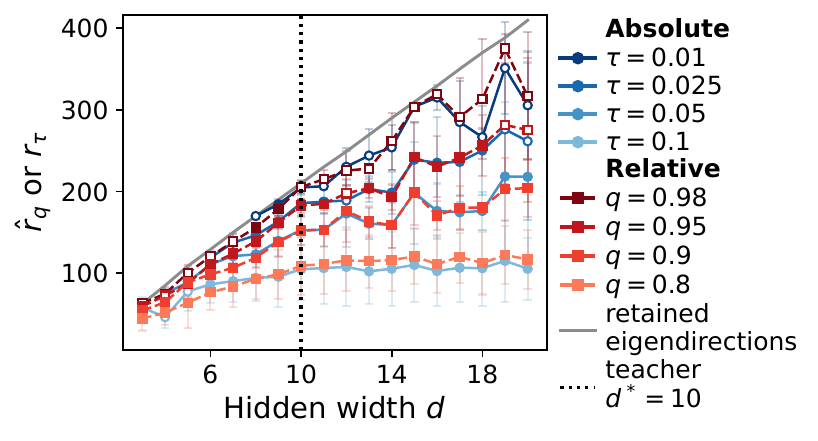}}
    \caption{Same setup as Fig.~\ref{fig:2regimes}(b) with tighter tolerances. Hollow markers indicate that fewer than five seeds reach the tolerance. The grey line shows the number of retained eigendirections, i.e.\ eigenpairs with $\lambda_j\geq10^{-6}\lambda_1$ (App.~\ref{app:implementation}).}
    \label{fig:tight}
\end{figure}

\textbf{Tighter tolerances.}
Fig.~\ref{fig:tight} repeats Fig.~\ref{fig:2regimes}(b) with tighter tolerances, without and with noise. For tight tolerances ($\tau=0.01$, $q=0.98$), the behavioral recovery rank keeps growing with the width and approaches the number of retained eigendirections. We attribute this to the fact that at such tolerances, many directions of small but nonzero curvature start to matter, and their number grows with the width. 
For some seeds, the tolerance is not reached within the retained eigendirections, i.e.\ those with $\lambda_j\geq10^{-6}\lambda_1$ (App.~\ref{app:implementation}), and the rank is undefined. These seeds would require directions with smaller eigenvalues, which we treat as part of the kernel and do not include. This means that the hollow markers can understate the rank.

\textbf{Quadratic approximation along the projection path.}

\begin{figure}[h]
    \centering
    \subfigure[Noiseless.]{
    \includegraphics[width=.4\textwidth]{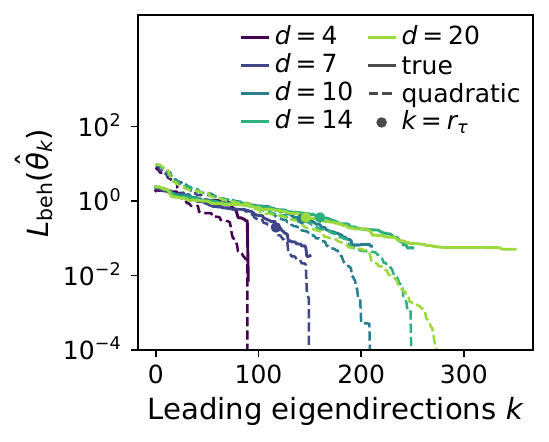}}
    \hfill
    \subfigure[$\sigma^2=0.01$.]{
    \includegraphics[width=.4\textwidth]{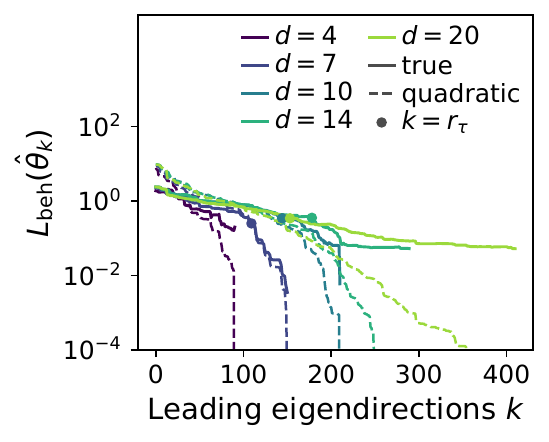}}
    \caption{Behavioral loss (solid) and its quadratic prediction from Eq.~\eqref{eq:residual_separable} (dashed) along the projection path for five student widths and one seed. The dots mark $r_\tau$ for $\tau=0.075$.}
    \label{fig:locality}
\end{figure}

The ordering of the eigendirections is motivated by the quadratic model of the behavioral loss. However, for small $k$, the projected parameters $\hat\theta_k$ can lie far from $\theta_\ast$. Fig.~\ref{fig:locality} compares the behavioral loss $L_{\mathrm{beh}}(\hat\theta_k;\theta_\ast,\mathcal P)$ with its quadratic prediction $Q_k=\tfrac12\sum_{j>k}\lambda_j\langle v_j,\theta_\ast-\theta_{\mathrm{ref}}\rangle^2$ from Eq.~\eqref{eq:residual_separable} along the projection path. It overestimates the loss for small $k$, is of the same order around $k=r_\tau$, and underestimates it for large $k$ in students wider than the teacher, where the component of $\theta_\ast$ outside the retained eigendirections is not restored.

\begin{figure}[h]
    \centering
    \subfigure[Rank versus width at five checkpoints.]{
    \includegraphics[width=.31\textwidth]{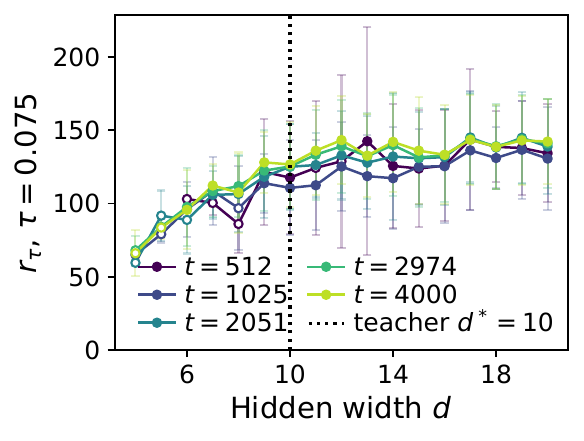}}
    \hfill
    \subfigure[Rank over training for selected widths.]{
    \includegraphics[width=.31\textwidth]{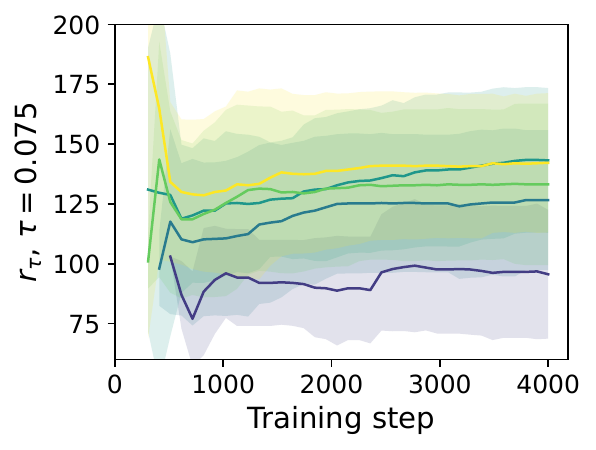}}
    \hfill
    \subfigure[Test MSE over training for the same widths.]{
    \includegraphics[width=.31\textwidth]{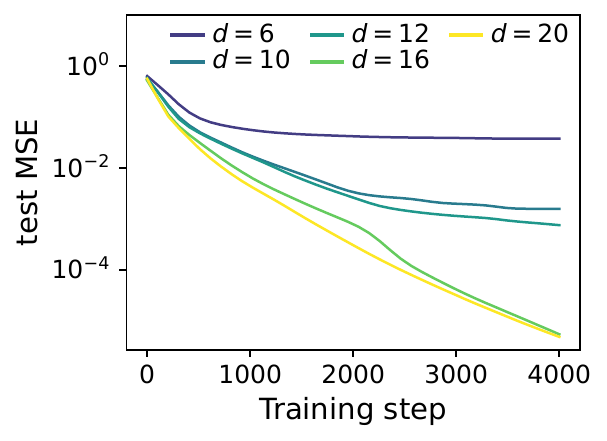}}
    \caption{Behavioral recovery rank $r_\tau$ for $\tau=0.075$ during training of the teacher-student models, mean and std.\ over the seeds that reach $\tau$. Hollow markers in (a) indicate that fewer than five seeds reach it. The colors in (b) correspond to the widths in the legend of (c).}
    \label{fig:temporal}
\end{figure}

\textbf{Recovery rank during training.}
Fig.~\ref{fig:temporal} shows the behavioral recovery rank for $\tau=0.075$ at $40$ checkpoints during the training of the teacher-student models. The tolerance is not attainable in the first few hundred steps and for the narrowest students. Once it is attainable, the dependence of the rank on the width has the same shape at every checkpoint (Fig.~\ref{fig:temporal}(a)). For students at least as wide as the teacher, the mean rank increases by at most $25$ directions between step $1025$ and the end of training (Fig.~\ref{fig:temporal}(b)), while the test MSE decreases by one to three orders of magnitude (Fig.~\ref{fig:temporal}(c)). The number of directions needed to reach a fixed tolerance is therefore reached early in training and changes little afterwards.

\clearpage\subsection{MNIST}

\begin{figure}[ht]
    \centering

    \subfigure[Test accuracy.]{
    \includegraphics[width=.38\textwidth]{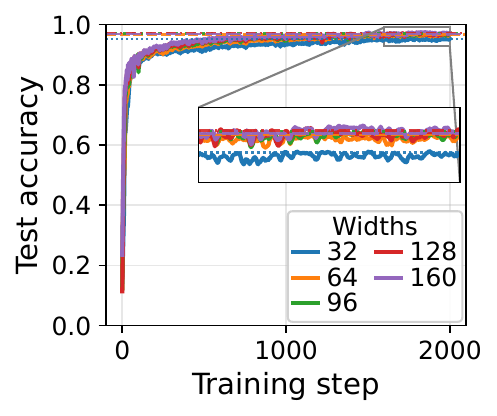}
    }
    \hspace{.8cm}
    \subfigure[Test loss.]{    \includegraphics[width=.38\textwidth]{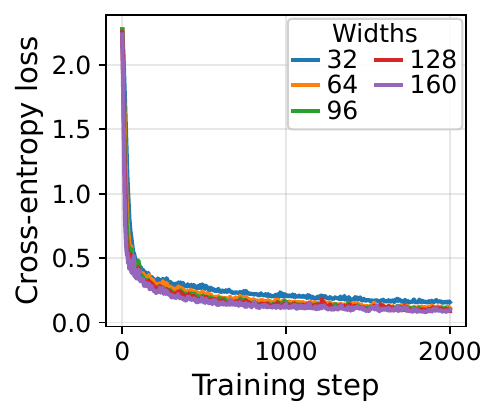}}

    \caption{For different widths of the MLP on MNIST, we report the accuracy and loss over the test set. The final accuracy exceeds $95\%$ for all widths. }
    \label{fig:mnist_evol}
\end{figure}

\textbf{Evolution over training.} Fig.~\ref{fig:mnist_evol} shows test loss and accuracy evolution over training, reaching final accuracies exceeding $.95, $ for all considered widths.

\begin{figure}[ht]
    \centering
    \subfigure[Training loss.]{  \includegraphics[width=0.38\textwidth]{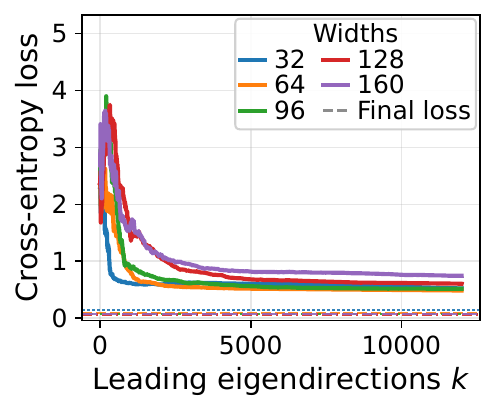}}
           \hspace{.8cm}
            \subfigure[Training accuracy.]{
            \includegraphics[width=0.38\textwidth]{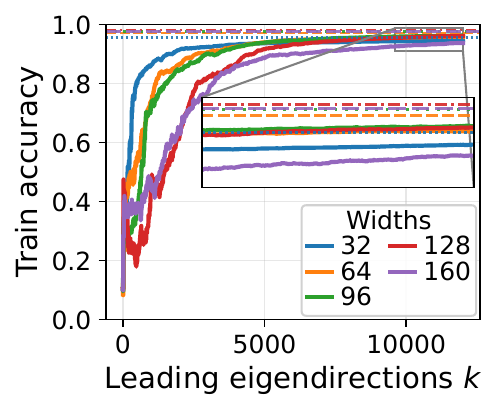}}
         
    \caption{Additional MNIST experiments, showing the training loss and training accuracy over the number of leading eigendirections.}
  \label{fig:mnist_additional}
\end{figure}

\textbf{Projection.} In Fig.~\ref{fig:mnist_additional}, we show the projection impact also on the training loss and accuracy.

\begin{figure}[t]
    \centering
     \subfigure[Eigenvalues over $k$, for different widths. ]{
    \includegraphics[height=5cm]{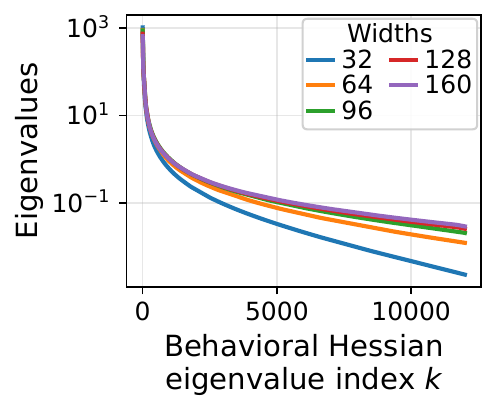}}
\hspace{.8cm}
              \subfigure[Number of eigenvalues needed to reach $q \times \sum_i^m \lambda_i$  among the top $m=12{,}000$ ones, for different widths. ]{    
    \includegraphics[height=5cm]{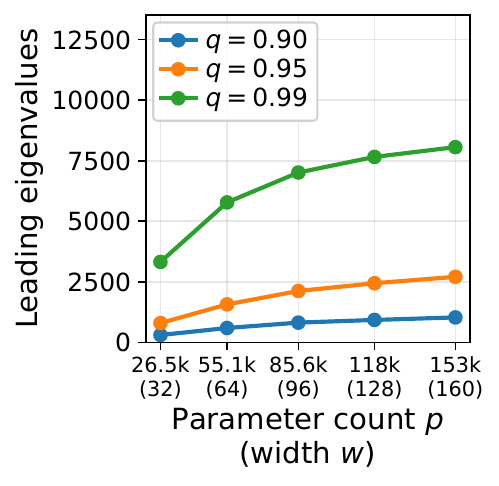}}
         \subfigure[Proportion of the top $k$ eigenvalues among the top $12{,}000$ ones, $ \sum_i^k \lambda_i/\sum_i^{12,000} \lambda_i$, for different widths. Dotted lines indicate the thresholds $q=0.90, 0.95, 0.99$.]{    
    \includegraphics[width=.38\textwidth]{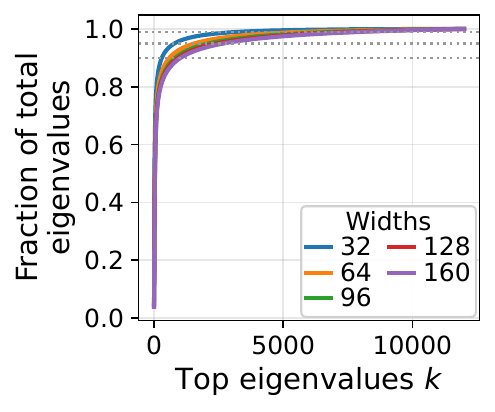}}
    \caption{Spectral quantities for the MNIST experiments.}
    \label{fig:mnist_additional_spectral}
\end{figure}

\textbf{Spectral structure.} Fig.~\ref{fig:mnist_additional_spectral} shows a  concentrated behavioral spectrum, further supporting the view that only a restricted subset of local parameter directions dominates functional recovery.

\end{document}